\pdfoutput=1

\documentclass[11pt]{article}
\usepackage[final]{acl}
\usepackage{times}
\usepackage{latexsym}

\usepackage{etoc}
\usepackage{etoolbox}

\usepackage[T1]{fontenc}

\usepackage[utf8]{inputenc}

\usepackage{microtype}

\usepackage{inconsolata}

\usepackage{graphicx}

\usepackage{kotex}

\usepackage{kotex-logo}
\usepackage{verbatim}
\usepackage{fixltx2e}
\usepackage{booktabs}
\usepackage{multirow}
\usepackage{multicol}
\usepackage{amsmath}
\usepackage{amssymb}
\usepackage{pifont} 

\usepackage[dvipsnames]{xcolor}
\usepackage{colortbl}
\usepackage{tablefootnote}
\usepackage[most]{tcolorbox}
\usepackage{wrapfig,boxedminipage,caption}
\usepackage[utf8]{inputenc}
\usepackage{array}
\usepackage{enumitem}

\usepackage{bbm}
\usepackage{longtable}
\usepackage{arydshln}
\usepackage{tabularx}
\usepackage[most]{tcolorbox} 
\usepackage{makecell}
\usepackage{mdframed}

\usepackage{float}  
\usepackage{capt-of} 
\usepackage{placeins}

\newfloat{example}{t}{lop}  
\floatname{example}{Example}  

\newcommand{\tact}{\includegraphics[height=0.9em]{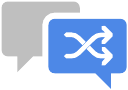}}

\newcommand{\cmark}{{\color{green!60!black}\ding{51}}}
\newcommand{\xmark}{{\color{red!70!black}\ding{55}}}

\definecolor{gold}{RGB}{255, 215, 0}
\definecolor{silver}{RGB}{192, 192, 192}
\definecolor{bronze}{RGB}{205, 127, 50}

\definecolor{og}{rgb}{0.42, 0.56, 0.28}
\definecolor{Gray}{gray}{0.95}
\newcolumntype{a}{>{\columncolor{Gray}}c}
\newcolumntype{b}{>{\columncolor{Gray}}l}
\newcolumntype{x}[1]{>{\raggedright\let\newline\\\arraybackslash\hspace{0pt}}m{#1}}
\newcolumntype{y}[1]{>{\raggedright\let\newline\\\arraybackslash\hspace{0pt}\columncolor{Gray}}m{#1}}

\title{Beyond Task-Oriented and Chitchat Dialogues: \\Proactive and Transition-Aware Conversational Agents}

\author{
  \textbf{Yejin Yoon}$^{\dagger}$ \quad
  \textbf{Yuri Son}$^{\dagger}$ \quad
  \textbf{Namyoung So}$^{\dagger}$ \quad
  \textbf{Minseo Kim}$^{\dagger}$ \\[0.5em]
  \textbf{Minsoo Cho}$^{\ddagger}$ \quad
  \textbf{Chanhee Park}$^{\ddagger}$ \quad
  \textbf{Seungshin Lee}$^{\ddagger}$ \quad
  \textbf{Taeuk Kim}$^{\dagger}$\thanks{Corresponding author} \\
  \\
  $^{\dagger}$Hanyang University \quad $^{\ddagger}$Hyundai Motor Company \\
  \texttt{\{stillwithyou,yurison,thskadud,er1123090,kimtaeuk\}@hanyang.ac.kr} \\
}

\begin{document}
\maketitle

\begin{abstract}


Conversational agents have traditionally been developed for either task-oriented dialogue (TOD) or open-ended chitchat, with limited progress in unifying the two. 
Yet, real-world conversations naturally involve fluid transitions between these modes. 
To address this gap, we introduce \tact\ \textbf{TACT} (\textbf{T}OD-\textbf{A}nd-\textbf{C}hitchat \textbf{T}ransition), a dataset designed for transition-aware dialogue modeling that incorporates structurally diverse and integrated mode flows. 
TACT supports both user- and agent-driven mode switches, enabling robust modeling of complex conversational dynamics.
To evaluate an agent’s ability to initiate and recover from mode transitions, we propose two new metrics---Switch and Recovery.
Models trained on TACT outperform baselines in both intent detection and mode transition handling. 
Moreover, applying Direct Preference Optimization (DPO) to TACT-trained models yields additional
gains, achieving 75.74\% joint mode-intent accuracy and a 70.1\% win rate against \textsc{GPT-4o} in human evaluation.
These results demonstrate that pairing structurally diverse data with DPO enhances response quality and transition control, paving the way for more proactive and transition-aware conversational agents.

\end{abstract}

\section{Introduction}



Conversational agents are generally classified into two types: task-oriented dialogue (TOD) systems and chitchat models for open-domain social interaction. 
TOD systems \cite{hosseini2020simple} follow predefined workflows to accomplish user goals, while chitchat models \cite{wu-yan-2018-deep} generate contextually appropriate responses.

\begin{figure}[!ht]
    \begin{center}
        \includegraphics[width=\linewidth]{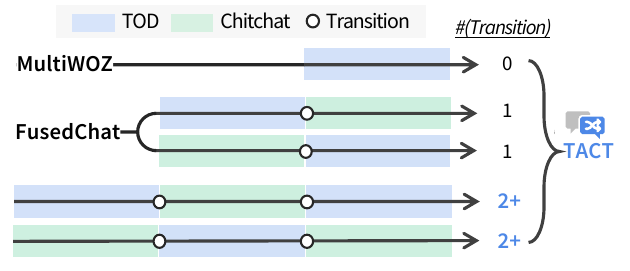} 
        \caption{
        Comparison of dialogue flows across MultiWOZ, FusedChat, and \tact\ TACT.
        MultiWOZ contains only task-oriented dialogues (TODs) with no mode transitions, whereas FusedChat allows a single transition between TOD and chitchat. In contrast, \tact\ TACT supports multiple mode transitions (2+), enabling training on complex dialogues with diverse switching patterns.
        }
        \label{fig:dialogue_flow}
    \end{center}
\end{figure}

With the aid of large language models (LLMs), research bridging the two traditionally independent paradigms has gained momentum \cite{sekulic-etal-2024-reliable}. 
However, current efforts overlook a core challenge: most dialogue systems remain predominantly \textit{reactive}, focusing on responding to user inputs rather than \textit{proactively} coordinating dialogue flows \cite{yi2024survey, acikgoz2025desideratum}.

\begin{figure*}[t]
    \begin{center}
        \includegraphics[width=\linewidth]{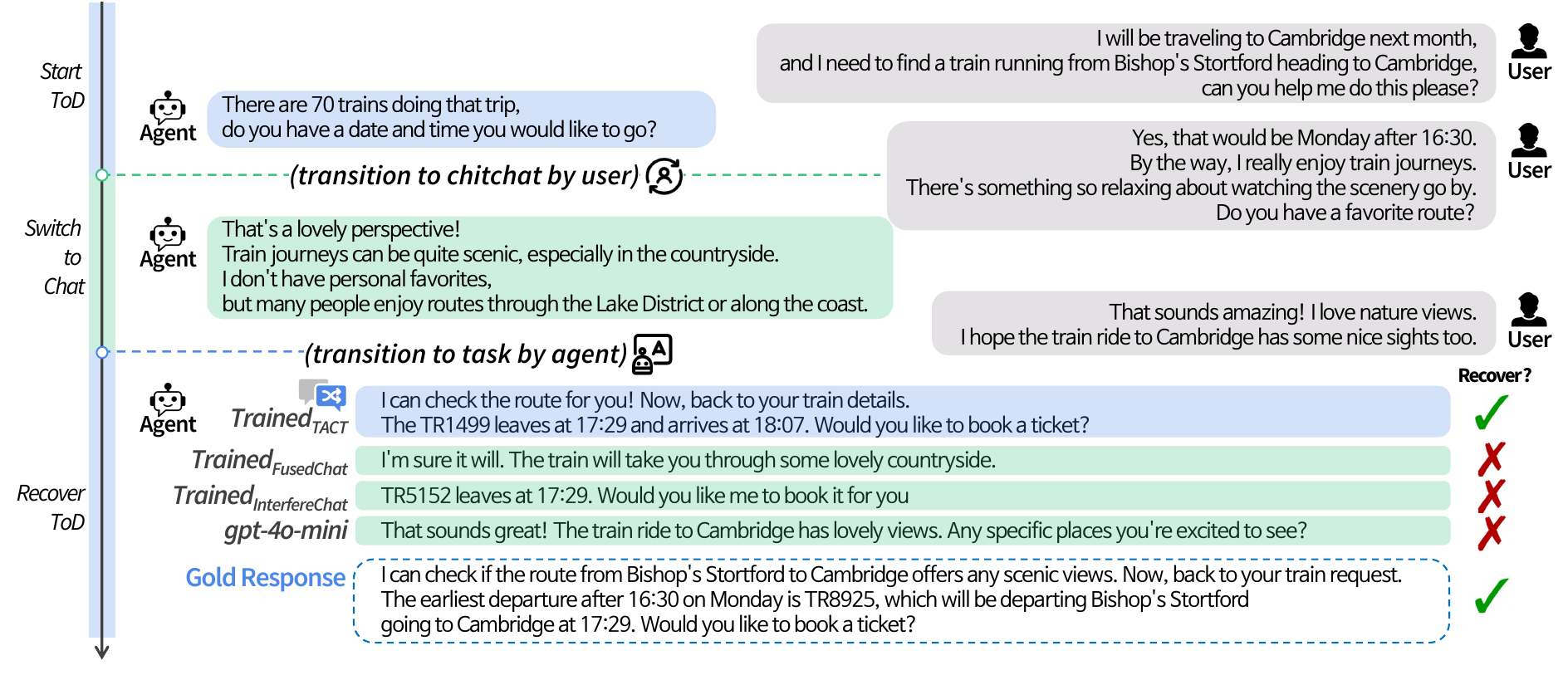}
        \caption{
        A scenario demonstrating that only \tact\ \textbf{Agent-$\text{Trained}_{\text{TACT}}$} exhibits transition-awareness and proactivity by successfully returning to the task after a chitchat. Note that all other baselines fail to recover the original context.
        }
        \label{fig:transition_comparison}
    \end{center}
\end{figure*}

In real-world conversations, users frequently shift between task-oriented dialogue (TOD) and chitchat within a single session, as empirically observed in deployed systems \citep{rim-etal-2025-chat}.\footnote{For example, during a conversation with a ticket-booking agent, a user might share past travel experiences. Ideally, the agent should respond appropriately to the digression, then autonomously return to the booking task when appropriate to fulfill its objective. See Figure \ref{fig:transition_comparison} for a similar case study.} 
To manage these dynamic transitions, agents must predict mode shifts, take initiative, and maintain coherent multi-turn interactions.
This requires two key abilities: (1) \textit{transition-awareness}, for detecting and adapting to mode changes, (2) \textit{proactivity}, to plan ahead and guide the conversation flow when appropriate---both of which are essential yet underexplored \cite{yi2024survey, acikgoz-etal-2024-bridging}.

In this work, we present a framework for building \textit{proactive} and \textit{transition-aware} conversational agents, grounded in our novel \tact\ \textbf{TACT}\footnote{Available at \url{https://github.com/HYU-NLP/TACT}.} (\textbf{T}OD-\textbf{A}nd-\textbf{C}hitchat \textbf{T}ransition) dataset. 
As illustrated in Figure~\ref{fig:dialogue_flow}, TACT features complex dialogue sessions in which task-oriented and chitchat modes frequently alternate. 
In contrast to prior mode-switching datasets \citep{chang-chen-2024-injecting, Young_Xing_Pandelea_Ni_Cambria_2022, stricker-paroubek-2024-chitchat}, which lack structural diversity (see Table \ref{tab:dataset_stat}\footnote{We treat InterfereChat solely as T(OD), as chitchat utterances always coexist with TOD turns. We also re-define the intent space of SLURP to eliminate overlaps between intents. We refer readers to Table \ref{tab:intent_flow_full} for more details.}),  TACT provides richer interactions that better support the training of agents to detect and manage mode shifts with both transition-awareness and proactivity. 
For instance, Figure~\ref{fig:transition_comparison} showcases a scenario in which only the TACT-based agent successfully returns to the original task after an interruption, whereas other models fail to recover and remain off track.

Furthermore, we define metrics to quantify how often agents attempt to switch or recover dialogue modes, and whether those transitions succeed based on user responses. 
Compared to previous datasets, agents trained on TACT consistently handle flow switching and recovery more effectively, while maintaining strong performance on standard TOD and chitchat tasks.
Beyond structural coverage, we further adopt Direct Preference Optimization (DPO; \citet{rafailov2023direct}) to align model outputs with human preferences, which significantly enhances response quality and transition naturalness.
These results suggest the potential to develop more autonomous and predictive conversational agents, surpassing the current standard where mere response accuracy is considered sufficient.

\section{Related Work}

\begin{table}[t]
    \centering
    \scriptsize
    \setlength{\tabcolsep}{0.5pt}
    \begin{tabular}{lcccaa}
        \toprule
        \textbf{Dataset} & \tiny{\textbf{SalesBot2.0}} & \tiny{\textbf{FusedChat}} & \tiny{\textbf{InterfereChat}} & 
        \multicolumn{2}{a}{\tiny{\tact\ \textbf{TACT}}} \\
        \midrule
            \textbf{Seed} & \tiny{SalesBot1.0} & \tiny{MultiWOZ2.4} & \tiny{FusedChat} & \tiny{MultiWOZ2.2} & \tiny{SLURP} \\
        \midrule
            \textbf{\# Intents} & 6 & 11 & 11 & 11 & 50* \\
            \textbf{\# Dialog} & 5,453 & 10,436 & 4,475 & 7,199 & 9,936 \\
            \textbf{\# Avg. Turn} & 7.71 & 18.36 & 13.58 & 15.04 & 16.42  \\
            \textbf{\# Avg. Switch} & 0.96 & 1 & 0* & 1.93 & 2.06 \\
            \textbf{\# Avg. Recov.} & - & 0 & 0* & 0.93 & 1.07 \\
        \midrule
            \textbf{\# Uniq. Flow} & 2 & 2 & 1 & 11 & 12 \\
            \textbf{Flow Types} & CT & TC, CT & T* & \multicolumn{2}{a}{TCT, CTC, TCTCT, etc.}\\
        \bottomrule
    \end{tabular}
    \caption{
    Statistics of existing datasets and the newly proposed \tact\ TACT variants. 
    Asterisks (*) indicate factors affected by the pre-processing performed in this work.
    \textbf{Recov.}: recovery.
    \textbf{\# Uniq. Flow}: the number of unique dialogue mode transition patterns.
    }
\label{tab:dataset_stat}
\end{table}

\paragraph{Datasets for dialogue mode switching}

Previous efforts to unify TOD and chitchat have largely relied on augmenting existing TOD corpora. 
FusedChat \cite{Young_Xing_Pandelea_Ni_Cambria_2022} and InterfereChat \cite{stricker-paroubek-2024-chitchat} inject chitchat turns into TOD, typically at fixed points or as single exchanges, hereby adhering to a TOD-centric perspective.
Such simple rule-based modification restricts TOD-chitchat variations, rendering these resources unsuitable for modeling dynamic dialogue transitions. To overcome this shortcoming and support rich transitions, we construct two TACT variants based on SLURP \cite{bastianelli-etal-2020-slurp} and MultiWOZ 2.2 \cite{zang-etal-2020-multiwoz}, featuring diverse intents and interwoven mode switches. 

\paragraph{Methods for dialogue mode switching}

While prior research, e.g., SimpleTOD \cite{hosseini2020simple} and SalesAgent~\cite{chang-chen-2024-injecting}, have made progress in unifying TOD and chitchat, these methods fall short in managing fluent mode control across multi-turn interactions.
Recent work has extended these efforts by integrating TOD and chitchat systems (Accentor; \citet{sun2020adding}) or employing function calling \cite{stricker2024few, li2024large} for accurate intent tracking. 
However, a critical aspect remains underexplored: the ability to recover and re-engage with relevant dialogue history at appropriate turns. 

\paragraph{Programmable control}
Prior work on controllable dialogue frameworks \cite{rebedea-etal-2023-nemo} enables rule-driven flow control, effective when conversation graphs are known in advance.
However, while these systems achieve tight control over dialogue flows---akin to classic pipelines \cite{bocklisch2017rasa, liu2021benchmarking}---they are inherently rigid and costly to adapt to new scenarios. 
In contrast, we address more challenging and practical settings where dialogues shift dynamically between chitchat and tasks without rigid scripts. 
To this end, we frame transition handling as a learning problem and construct TACT to capture such phenomena without predefined flows. 
Building on this, we propose a unified framework that fine-tunes LLMs with TACT for proactive transition and flow control, while incorporating response-level preferences through DPO.

\section{Dataset Construction: \tact\ TACT}
\label{sec:tact_construction}

In this section, we design the \textbf{TACT} (\textbf{T}OD-\textbf{A}nd-\textbf{C}hitchat \textbf{T}ransition) dataset to support agents that manage both TOD and chitchat, through coherent multi-turn dialogues. 
Unlike prior datasets, it regards mode transitions as recoverable and context-sensitive, allowing agents to decide when to switch and how to resume back to previous modes. 
We release the dataset as open-source to support future research in multi-turn dialogue systems.\footnote{\url{https://huggingface.co/datasets/HYU-NLP/TACT}.}




\subsection{Dialogue Generation} \label{sec:dialogue_generation}

\begin{figure*}[t]
    \begin{center}
        \includegraphics[scale=0.55]{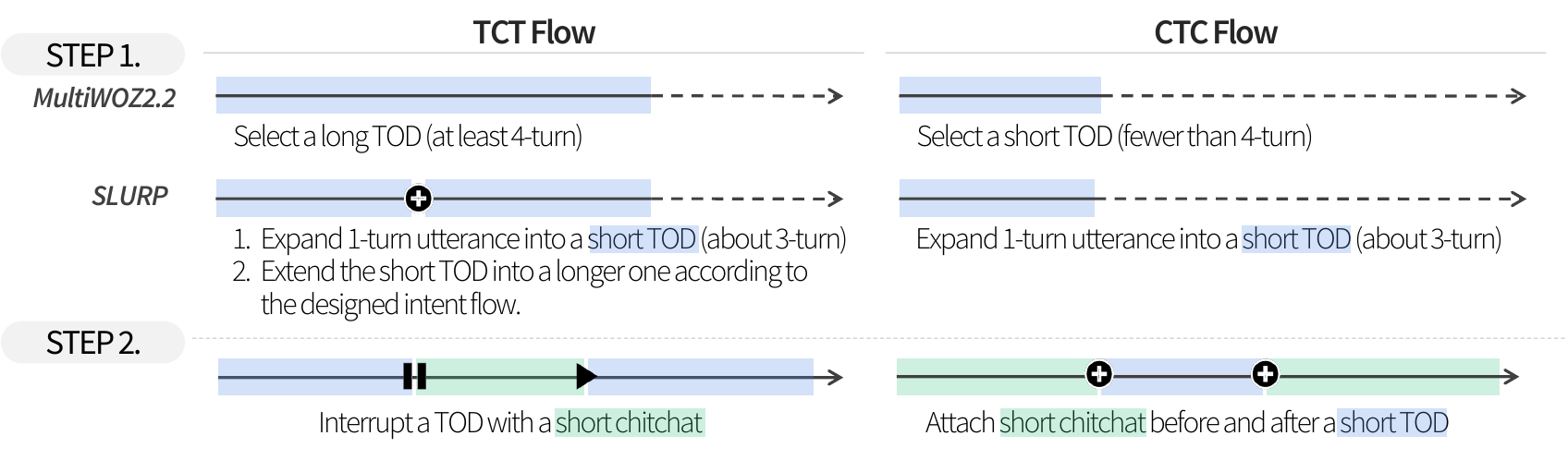} 
        \caption{ \tact\ TACT construction steps for TCT and CTC dialogues from MultiWOZ and SLURP.
        }
        \label{fig:dialogue_generation}
    \end{center}
\end{figure*}

To support structured modeling of mode transitions, \tact\ TACT defines two core dialogue flow types: \textbf{TCT} (TOD $\mapsto$ Chitchat $\mapsto$ TOD) and \textbf{CTC} (Chitchat $\mapsto$ TOD $\mapsto$ Chitchat). 
These flow types simulate patterns observed in real-world conversations, where users often deviate from a task and later return to it, or casually initiate and conclude chitchat around brief task-oriented exchanges.

We construct such flows based on existing TOD corpora—MultiWOZ2.2 and SLURP
---by augmenting them with chitchat in contextually appropriate locations (see Figure \ref{fig:dialogue_generation} and Appendix \ref{sec:appendix_tact_construction} for details). 
This approach supports modeling of both transition awareness and recovery, which are critical for fluid, mode-integrated dialogue.

\paragraph{TCT flow}

We extract task segments of four or more turns from MultiWOZ2.2, and insert a chitchat block at a natural boundary between intents. 
The chitchat briefly diverges from the task, often reflecting personal curiosities or preferences, before the dialogue returns to its original goal.

\paragraph{CTC flow}

We begin with a short TOD segment (2–3 turns) and attach chitchat before and after the task, forming a wrap-around flow. 
This simulates cases where users casually engage in a task during a social exchange and then resume chitchat.

\subsection{Dataset Validation} \label{sec:validation}


\begin{table}[t]
    \centering
    \scriptsize
    \setlength{\tabcolsep}{0.3em}
    \begin{tabular}{lccc}
        \toprule
        \textbf{\makecell[l]{Validation\\Approaches}}
        & \textbf{\makecell{Human-Annotated\\Criteria}}
        & \textbf{\makecell{LLM-Generated\\Task Description}}
        & \textbf{\makecell{LLM-Generated\\Evaluation Steps}} \\
        \midrule
        \textbf{Active Critic} & \xmark & \cmark & \xmark \\
        \textbf{G-Eval} & \cmark & \xmark & \cmark \\
        \textbf{Ours} & \cmark & \cmark & \cmark \\
        \bottomrule
    \end{tabular}
    \caption{
    Comparison of data validation methods. 
    Only our framework supports both human-aligned criteria and LLM reasoning.
    \cmark\ indicates feature presence.
    }
    \label{tab:validation_comparison}
\end{table}

To ensure the quality of TACT at scale, we develop an automatic validation pipeline fusing human-aligned criteria with model-based reasoning.
Specifically, G-Eval \cite{liu-etal-2023-g} offers well-designed evaluation criteria aligned with human judgment, but lacks internal reasoning. 
In contrast, Active-Critic \cite{xu2024large} induces task-specific criteria through reasoning over examples, but it is not designed for dialogue evaluation and does not use predefined human-authored standards.
Table~\ref{tab:validation_comparison} shows that our method uniquely combines human-authored criteria with model-based reasoning.\footnote{Full prompt templates are provided in Figure~\ref{fig:appendix_prompt_validation}.}
Unlike G-Eval or Active-Critic, which cover only one of these aspects, our hybrid framework supports scalable and interpretable validation.

Dialogues are evaluated with a prompt that includes a full conversation, a task description, evaluation criteria, and step-by-step reasoning. 
The judging model, \textsc{GPT‑4o-mini} \citep{hurst2024gpt}, scores each dialogue on three criteria:
(1) \textbf{Intent Accuracy}, which checks whether the user intent is correctly conveyed and executed;
(2) \textbf{Transition Quality}, which assesses whether mode switches are contextually justified; and
(3) \textbf{Dialogue Naturalness}, which evaluates fluency and coherence.

\subsection{Dataset Characteristics} 
\label{sec:dataset_analysis}

\begin{figure}[t]
    \begin{center}
        \includegraphics[width=0.97\linewidth]{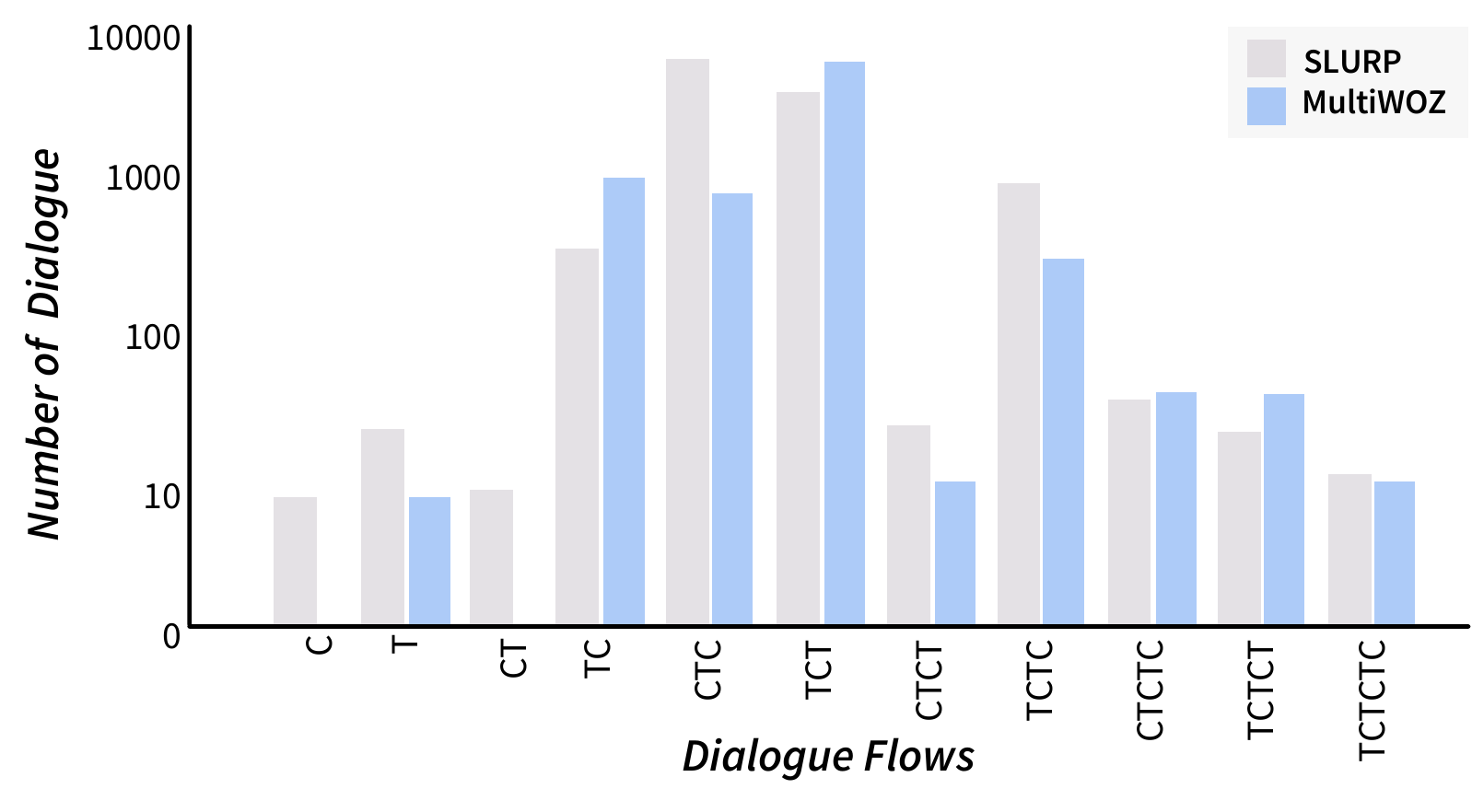} 
        \caption{
        Dialogue flow distribution in TACT$_\text{SLURP}$ and TACT$_\text{MultiWOZ}$. 
        }
        \label{fig:flow_histogram}
    \end{center}
\end{figure}

\tact\ TACT is structurally designed to support robust learning of dialogue transitions through two key properties: diverse multi-turn flow patterns and recoverable dialogue structures.


\paragraph{Diverse transition flows} 
As illustrated in Table~\ref{tab:dataset_stat}, TACT supports a wide variety of dialogue flows, including TCT, CTC, TCTCT, and others. 
TACT is the first dataset to combine interwoven transition patterns, a balanced distribution of transition initiators, and recoverable dialogue structures.
Figure~\ref{fig:flow_histogram} reveals that while TACT$_\text{MultiWOZ}$ examples concentrate more on TCT and TC flows, TACT$_\text{SLURP}$ covers a broader spectrum, enabling agents to generalize across diverse flow structures.

\paragraph{Recoverable structures}
Differing from existing datasets, TACT includes dialogues that explicitly return to a previously suspended mode, enabling agents to learn dialogue-level consistency. 
For example, Figure~\ref{fig:transition_comparison} shows a TCT dialogue where only the \tact\ TACT-trained model resumes the original task after a chitchat interruption.
We provide a detailed evaluation of recovery performance in \S ~\ref{sec:discussion}.




\section{Methodology} 
\label{sec:methodology}

We evaluate 4 methods for unified TOD-chitchat response generation:
(1) \textbf{in-context learning (ICL)},
(2) \textbf{supervised fine-tuning (SFT)}
(3) \textbf{Direct Preference Optimization (DPO)}, our main approach, which aligns model outputs with response preferences, and
(4) \textbf{pipeline-based methods} that separate mode prediction from response generation.


\begin{table*}[h]
    \small
    \centering
    \resizebox{\textwidth}{!}{%
    \begin{tabular}{ccaaccaaccaa}
        \toprule
        & & \multicolumn{6}{>{\cellcolor{blue!10}}c}{\textbf{TOD}} & \multicolumn{4}{>{\cellcolor{yellow!30}}c}{\textbf{Flow}}  \\
        \arrayrulecolor{gray!70} \cmidrule(){3-12} \arrayrulecolor{black}
        {\textbf{ }} & \textbf{} & \multicolumn{2}{a}{\textbf{Mode Selection}} & \multicolumn{2}{c}{\textbf{Intent Detection}} & \multicolumn{2}{a}{\textbf{Joint Accuracy}} &\multicolumn{2}{c}{\textbf{Switch}} & \multicolumn{2}{a}{\textbf{Recovery}} \\
        $\textbf{Training Set}$ & \textbf{Test Set} & \textbf{Acc.} & \textbf{F1} & \textbf{Acc.}/turn & \textbf{Acc.}/dialogue & \textbf{Acc.}/turn & \textbf{Acc.}/dialogue & \textbf{Attempt} & \textbf{Success} & \textbf{Attempt} & \textbf{Success} \\
        \midrule
        \multirow{5}{*}{$\text{FusedChat}$} 
            & MultiWOZ   & 98.44 & 76.71 & 93.79 & 69.80 & 93.57 & 68.90 & 0.000 & 0.000 & - & - \\
            & FusedChat     & 97.00 & 96.70 & 92.85 & 66.90 & 94.20 & 60.50 & 0.000 & 0.000 & - & - \\
            & InterfereChat & 97.04 & 94.68 & 93.21 & 67.62 & 93.09 & 61.89 & 0.000 & 0.000 & - & - \\
            & $\text{TACT}_{\text{MultiWOZ}}$  & 91.79 & 87.27 & 94.46 & 72.65 & 88.13 & 33.24 & 0.000 & 0.000 & - & - \\
            \arrayrulecolor{gray!70} \cmidrule(l){2-8} \arrayrulecolor{black}
            & \textbf{Average} & \cellcolor{blue!20}96.07 & \cellcolor{blue!20}88.84 & \cellcolor{blue!20}93.58 & \cellcolor{blue!10}69.24 & \cellcolor{blue!30}\textbf{92.25} & \cellcolor{blue!20}56.13 &  \\
        \midrule
        \multirow{5}{*}{$\text{InterfereChat}$} 
            & MultiWOZ   & 98.27 & 75.97 & 93.92 & 70.90 & 93.74 & 70.10 & 0.000 & 0.000 & - & - \\
            & FusedChat     & 79.92 & 73.43 & 92.85 & 67.10 & 76.19 & 7.10 & 0.000 & 0.000 & - & - \\
            & InterfereChat & 97.63 & 95.72 & 93.28 & 68.44 & 93.34 & 64.34 & 0.000 & 0.000 & - & - \\
            & $\text{TACT}_{\text{MultiWOZ}}$  & 79.26 & 58.41 & 93.95 & 71.34 & 84.33 & 34.89 & 0.000 & 0.000 & - & - \\
            \arrayrulecolor{gray!70} \cmidrule(l){2-8} \arrayrulecolor{black}
            & \textbf{Average} & \cellcolor{blue!10}88.77 & \cellcolor{blue!10}75.88 & \cellcolor{blue!30}\textbf{93.84} & \cellcolor{blue!30}\textbf{70.62} & \cellcolor{blue!10}84.82 & \cellcolor{blue!10}35.39 &    \\
        \midrule
        \multirow{5}{*}{\tact\ ${\text{TACT}_{\text{MultiWOZ}}}$} 
            & MultiWOZ   & 98.06 & 74.91 & 92.70 & 66.20 & 92.57 & 65.50 & 0.000 & 0.000 & 1.000 & $<$ 0.001 \\
            & FusedChat     & 90.63 & 89.05 & 92.57 & 65.70 & 87.08 & 34.10 & 0.160 & 0.008 & 1.000 & $<$ 0.001 \\
            & InterfereChat & 97.32 & 95.20 & 92.13 & 64.14 & 92.38 & 59.22 & 0.619 & 0.309 & 0.013 & 0.104  \\
            & $\text{TACT}_{\text{MultiWOZ}}$  & 98.95 & 98.50 & 96.35 & 80.94 & 96.41 & 75.59 & 1.322 & 1.300 & 0.977 & 0.856  \\
            \arrayrulecolor{gray!70} \cmidrule(l){2-8} \arrayrulecolor{black}
            & \textbf{Average} & \cellcolor{blue!30}\textbf{96.24} & \cellcolor{blue!30}\textbf{89.42} & \cellcolor{blue!10}93.44 & \cellcolor{blue!20}69.25 & \cellcolor{blue!20}92.11 & \cellcolor{blue!30}\textbf{58.60} &  \\
        \bottomrule 
    \end{tabular}
    }
    \caption{
    Comparison of SFT variants trained on different datasets. The \tact\ TACT-based agent performs competitively across diverse metrics.
    The best average score in each column is in \textbf{bold}; `$<$ 0.001' indicates a value less than 0.001.
    }
    \label{tab:dataset_comparison}
\end{table*}

\begin{figure}[t]
    \begin{center}
        \includegraphics[width=\linewidth]{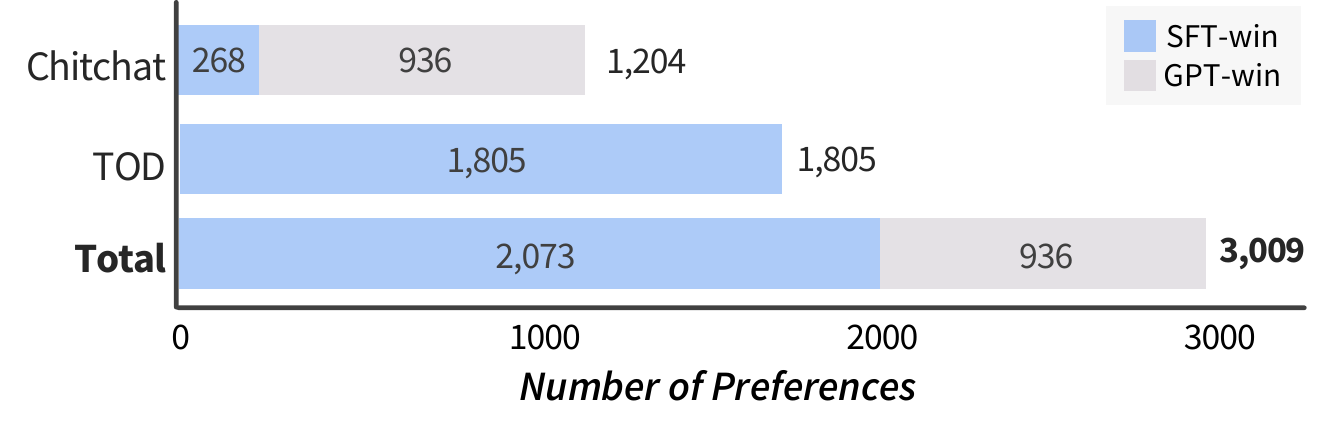} 
        \caption{
        Distribution of response preferences between the SFT and GPT models, evaluated by \textsc{Gemini-2.5-Pro} on TACT$_\text{MultiWOZ}$. For TOD dialogues, SFT-generated responses are consistently favored, whereas chitchat reveals a more mixed preference landscape.
        %
        }
        \label{fig:dialogue_mode_distribution}
    \end{center}
\end{figure}

\subsection{In-Context Learning}
\label{subsec:icl}

We use \textsc{GPT-4o} \citep{hurst2024gpt} to explore the potential of ICL for unified TOD-chitchat dialogue modeling, on both zero-shot and few-shot prompting setups. 
Prompt formats and example inputs are explained in Appendix~\ref{sec:appendix_icl_prompts}.
In the \textbf{zero-shot} setting, the model receives a task instruction and dialogue history as input, and is asked to predict both the current mode and corresponding responses.
In the \textbf{few-shot} setting, we provide four annotated exemplars: two for TCT flows and two for CTC. 
\subsection{End-to-End Generation with Finetuning}
\label{subsec:endtoend}

As our SFT baseline, we adopt \textbf{FnCTOD} \cite{li-etal-2024-large-language-models}, originally developed for zero-shot dialogue state tracking in unified TOD-chitchat scenarios using structured function calls.\footnote{
Among several candidate architectures---including SimpleToD \citep{hosseini2020simple} and SalesAgent \citep{chang-chen-2024-injecting}---FnCTOD demonstrated the best overall performance in our preliminary experiments (see Appendix \ref{sec:appendix_preliminary_eval}) and was thus selected as the main SFT approach.}
We reinterpret the function-calling mechanism as a structured intent representation: at each dialogue turn, the model first predicts an intent based on the user input, then produces a response conditioned on that intent.
The model is trained on \tact\ TACT using system-provided function schemas in the prompt, enabling unified intent prediction and response generation within a single auto-regressive decoding process.

\subsection{DPO for Unified TOD-Chitchat Modeling} \label{sec:dpo}


To improve both response quality and transition handling in hybrid dialogues, we further apply \textbf{DPO} \citep{rafailov2023direct} to the FnCTOD model trained on \tact\ TACT.
While SFT enables basic task completion, we observed its clear limitations in sensibleness, tone, and flow continuity, as reflected in the win rate outcomes of Figure~\ref{fig:generated_winrate}. 
Such shortcomings are consistent with recent findings \cite{chu2025sft} that SFT tends to emphasize memorization over generalization, whereas RL-based methods encourage more robust adaptation.

To this end, we apply DPO as a way to steer the model toward more desirable behaviors. 
By distinguishing between preferred and non-preferred responses, DPO guides the model toward accurate and complete outputs in TOD, while encouraging natural tone, continuity, and fluency in chitchat.
In this setting, \tact\ TACT exposes the model to dialogues where transitional contexts are clearly present and preferences can be readily determined.





Each training instance consists of an input prompt and two candidate responses: one \textit{preferred} and one \textit{rejected}.
We generate these pairs by comparing outputs from the FnCTOD and \textsc{GPT-4o-mini} models, with preferences determined by the \textsc{Gemini-2.5-Pro} \citep{gemini2023} judge according to the criteria in \S\ref{sec:chitchat-metrics}---i.e., sensibleness, specificity, interestingness, and transition naturalness.


The final dataset consists of 3,009 preference-labeled pair instances, as summarized in Figure~\ref{fig:dialogue_mode_distribution}.
Training on this dataset with DPO equips the model to balance precise task intent prediction with fluid, engaging conversational behavior.
We refer to the resulting model as \textbf{(SFT-)DPO}.
To the best of our knowledge, this is the first application of DPO in a unified dialogue generation setting that combines structured function-calling with TOD and chitchat.

\subsection{Pipeline Approach} 


Previous work on unified TOD–chitchat modeling typically adopts a modular pipeline: a classifier selects the dialogue mode, followed by a dedicated TOD or chitchat module for response generation \cite{Young_Xing_Pandelea_Ni_Cambria_2022}. 
To facilitate comparison with this paradigm, we construct a \textbf{generative-classifier-based pipeline}.\footnote{We also attempted to implement a variant with a BERT-based mode detector. However, this approach performed poorly in mode selection and exhibited low transition fluency.
We refer readers to Appendix \ref{sec:appendix_preliminary_eval} for more details.}
The FnCTOD model trained on TACT predicts a mode label and an intent. If the mode is TOD, it generates the response; otherwise, \textsc{GPT-4o-mini} is used with in-context prompts to generate a chitchat response.

\begin{table*}
    \small
    \centering
    \resizebox{\textwidth}{!}{%
    \begin{tabular}{caaccaaccaac}
        \toprule
        & \multicolumn{6}{>{\cellcolor{blue!10}}c}{\textbf{TOD}} & \multicolumn{4}{>{\cellcolor{yellow!30}}c}{\textbf{Flow}} & \cellcolor{green!10}\textbf{Chitchat} \\
        \arrayrulecolor{gray!70} \cmidrule(){2-12} \arrayrulecolor{black}
        \textbf{ } & \multicolumn{2}{a}{\textbf{Mode Selection}} & \multicolumn{2}{c}{\textbf{Intent Detection}} & \multicolumn{2}{a}{\textbf{Joint Accuracy}} &\multicolumn{2}{c}{\textbf{Switch}} & \multicolumn{2}{a}{\textbf{Recovery}} & \textbf{Overall} \\
        $\textbf{Method}$ &  \textbf{Acc.} & \textbf{F1-score} & \textbf{Acc.}/turn & \textbf{Acc.}/dialogue & \textbf{Acc.}/turn & \textbf{Acc.}/dialogue & \textbf{Attempt} & \textbf{Success} & \textbf{Attempt} & \textbf{Success} & \textbf{Win-Rate}\\
        \midrule
        ICL-ZS & 90.46 & 86.21 & 87.57 & 50.44 & 85.01 & 30.00 & 0.879 & 0.374 & 0.880 & 0.099 & - \\
        ICL-FS & 91.45 & 88.98 & 84.09 & 40.00 & 86.89 & 36.76 & \textbf{1.577} & 0.865 & \textbf{1.571} & 0.652 & - \\

        \midrule
        SFT & \textbf{98.95} & \textbf{98.50} & \textbf{96.35} & \textbf{80.94} & \textbf{96.41} & 75.59 & 1.322 & 1.300 & 0.977 & 0.856 & 23.16 \\
        SFT-DPO & 98.82 & 98.32 & 96.03 & 80.00 & 96.21 & \textbf{75.74} & 1.343& \textbf{1.322} & 0.977 & \textbf{0.859} & \textbf{40.86} \\
        \midrule
        Pipeline & \textbf{98.95} & \textbf{98.50} & \textbf{96.35} & \textbf{80.94} & \textbf{96.41} & 75.59 & 1.322 & 1.300 & 0.977 & 0.856 & 24.32 \\
        \bottomrule 
    \end{tabular}%
    }
    \caption{
    Method comparison across TOD, chitchat, and transition-aware metrics: ICL (zero- and few-shot with \textsc{GPT-4o}), SFT, (SFT-)DPO, and a generative-classifier-based Pipeline. 
    For Pipeline, TOD metrics are inherited from the SFT model.
    The best score in each column is in \textbf{bold}.
    }
    \label{tab:method_comparison}
\end{table*}

\section{Experimental Setup} 
\label{sec:experiment_setup}

\subsection{Datasets and Tasks}

We first investigate the performance of SFT-based model variants trained separately on four different datasets---\textbf{MultiWOZ2.2} \cite{zang-etal-2020-multiwoz}, \textbf{FusedChat} \cite{Young_Xing_Pandelea_Ni_Cambria_2022}, \textbf{InterfereChat} \cite{stricker-paroubek-2024-chitchat}, and our proposed \tact\ \textbf{TACT$_\text{MultiWOZ}$}\footnote{
Results on \textbf{TACT$_\text{SLURP}$} are presented in Appendix~\ref{sec:appendix_slurp}, showing similar trends with results discussed in \S \ref{sec:discussion}.}---for TOD–chitchat unification modeling.
The objective is to evaluate the impact of training data, demonstrating TACT’s effectiveness in activating transition-awareness and proactivity, while maintaining overall performance.

In the second part, we compare four modeling strategies to measure their relative effectiveness in terms of leveraging TACT:  
\textbf{ICL} using \textsc{GPT-4o} (zero- and few-shot),  
\textbf{SFT} with FnCTOD, \textbf{DPO}, and 
a \textbf{generative-classifier-based pipeline}.

\subsection{Training Configuration}
All SFT models, including DPO and pipeline variants, are initialized with \textsc{LLaMA-3.1-8B-Instruct} \citep{grattafiori2024llama} and fine-tuned using DeepSpeed ZeRO-3 \citep{deepspeed2021} with bf16 precision for efficiency and stability.  
We train each model for 3 epochs with a fixed learning rate of \(1 \times 10^{-5}\) and a batch size of 256.  
These hyper-parameters follow prior work, ensuring convergence and robust performance.
Implementation details are provided in Appendix~\ref{sec:appendix_env}.



\subsection{Evaluation Metrics} \label{sec:metrics}

\begin{figure}[t]
    \begin{center}
        \includegraphics[width=\linewidth]{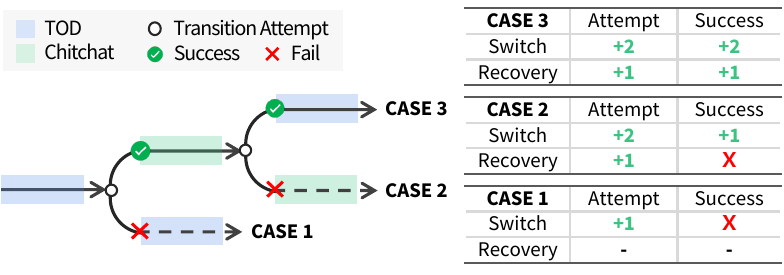} 
        \caption{
        Visualization of \textbf{Switch} and \textbf{Recovery} metrics.
        \textbf{Case 1}: Attempted Switch, but unsuccessful; \textbf{Case 2}: Successful switch but failed recovery; \textbf{Case 3}: Both switch and recovery succeed.
        }
        \label{fig:metric2}
    \end{center}
\end{figure}

Existing evaluation metrics for dialogue systems, e.g., slot accuracy, goal completion rate, and BLEU \cite{papineni2002bleu}, are mostly focused on TOD \cite{wen-etal-2017-network, RastogiZSGK20}.  
However, they are insufficient for considering scenarios involving frequent mode transitions between TOD and chitchat.  
A contemporary study \cite{acikgoz2025td} also highlights their inability to capture mid-dialogue failures and long-range inconsistencies.

To tackle this, we adopt a transition-aware evaluation framework with three categories:
TOD-centric measures, chitchat response quality (i.e., win rate), and transition-aware flow metrics.\footnote{Note that TOD-centric and chitchat response metrics are reported as percentages (w/o the \% symbol), while transition-aware measures are based on average attempts and successes.}

\paragraph{TOD-centric metrics}
We use three metrics to assess a model's task-handling abilities.
\textbf{(1) Mode Selection Accuracy and F1:}
Each turn is labeled as TOD or chitchat, and the model is evaluated for both overall accuracy and F1 scores.
\textbf{(2) Intent Detection Accuracy:}
For TOD-predicted turns, we evaluate whether the model accurately identifies the user’s intent.
Accuracy is reported at both the turn and dialogue level, where the latter requires perfect intent prediction across all TOD turns.
\textbf{(3) Mode+Intent Joint Accuracy:}
This metric checks if the dialogue mode and, when in TOD, the intent are correctly predicted at each turn, yielding end-to-end accuracy in integrated-mode settings.


\paragraph{Chat response quality metrics}\label{sec:chitchat-metrics}
We assess chat responses on four criteria: \textit{Sensibleness}, \textit{Specificity}, \textit{Interestingness} \citep{thoppilan2022lamda}, and     \textit{Transition Naturalness}, which evaluates the contextual appropriateness of mode switches.
For each criterion, we conduct pairwise comparisons between the target model and \textsc{GPT-4o} (few-shot), with judgments provided either by an LLM evaluator\footnote{To mitigate bias in model-based evaluation~\cite{li2025preference,wataoka2024self}, we employ \textsc{Gemini-2.5-Pro} as the judge, chosen for its independence from all evaluated models.} or by human annotators.
The pairwise win/lose outcomes are aggregated into win rates for each criterion.
To mitigate verbosity bias \cite{dubois2024length, hu2024explaining}, we apply length-controlled prompting for \textsc{GPT-4o} (see Appendix \ref{sec:appendix_response_quality}).

\paragraph{Transition-aware metrics}



To evaluate an agent's ability to manage mode transitions, we propose two transition-aware metrics:
\textbf{Switch}---when the agent shifts from one mode to another (e.g., TC, CT)---and \textbf{Recovery}---when the agent returns to a previously suspended mode (e.g., TCT, CTC).
For each metric, we report two statistics: 
(1) \textbf{Attempt}: the average number of \textit{Switch/Recovery} attempted \textit{\textbf{by the model}} per dialogue,   
(2) \textbf{Success}: the average number of \textit{Switch/Recovery} in which the user accepts the agent-driven mode shift and responds accordingly.
As shown in Figure \ref{fig:metric2}, Attempt is counted every time the agent suggests a possible mode transition, but it is only considered \textit{successful} when the user subsequently accepts the suggested mode transition. 
Formal definitions are provided in Appendix \ref{sec:appendix_metrics}.

\begin{figure*}[t]
    \centering
    \includegraphics[width=\linewidth]{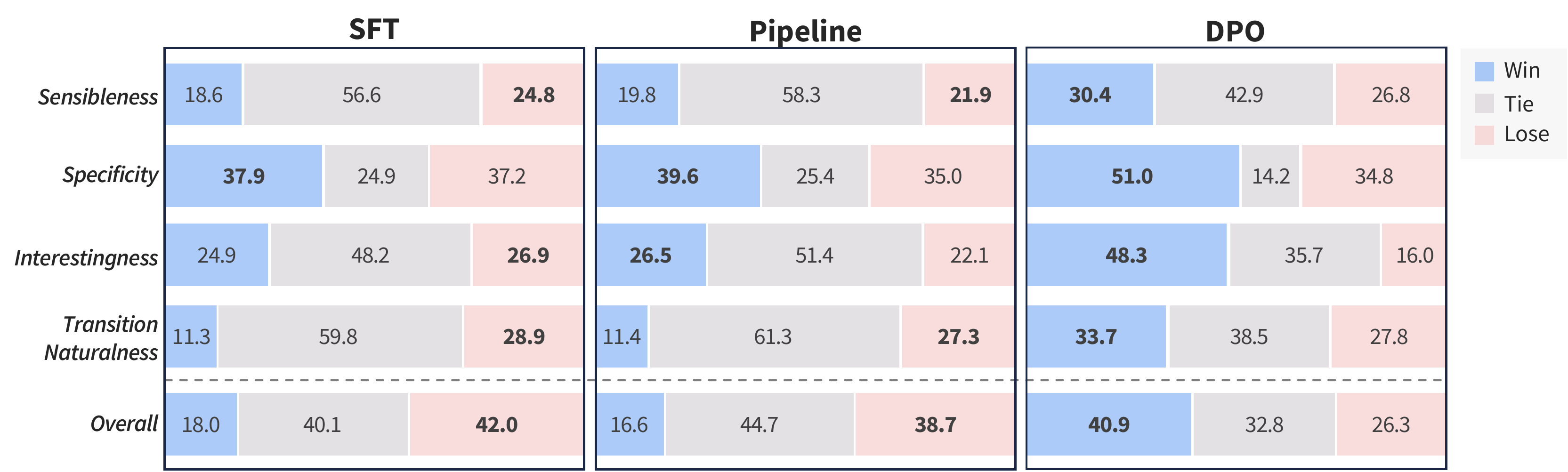}
    \caption{
    Preference-based evaluation of SFT, Pipeline, and DPO models (trained on \tact\ TACT) against \textsc{GPT-4o} (few-shot). 
    An LLM judge (\textsc{Gemini-2.5-Pro}) assessed sensibleness, specificity, interestingness, and transition naturalness, with win/tie/lose proportions shown.    
    }
    \label{fig:generated_winrate}
\end{figure*}


\section{Experimental Results}
\label{sec:discussion}




\subsection{Quantitative Evaluation} \label{sec:quantitative_results}



\paragraph{Comparison across datasets}
In Table~\ref{tab:dataset_comparison}, we compare SFT models trained on different TOD-chitchat unification datasets. 
The model trained on \tact\ TACT$_\text{MultiWOZ}$ achieves equal or superior performance on all TOD-centric metrics, compared to those trained on FusedChat or InterfereChat.

While FusedChat and InterfereChat-trained models perform well on their respective test sets, they generalize poorly to others.
In contrast, the TACT-trained one consistently achieves strong TOD performance and is the only one capable of handling multi-turn transitions and recoveries---a direct reflection of the dataset design described in \S\ref{sec:tact_construction}.


Notably, only the \tact\ TACT-trained agent achieves \textit{non-zero} transition-aware scores.
Other datasets produce no valid attempts, due to their lack of recoverable or multi-turn transition structures.  

\paragraph{Comparison across methods}
We compare four methods for handling dialogue with transitions: zero-  and few-shot prompting, SFT, a transition-aware pipeline, and preference-tuned DPO.
Table~\ref{tab:method_comparison} reveals that both SFT and the pipeline outperform ICL on TOD-centric metrics, especially in intent detection and mode+intent joint accuracy.  
While DPO slightly underperforms on these metrics, it achieves the highest dialogue-level joint accuracy at 75.74\%, suggesting greater consistency and better alignment with user intent.




DPO also outperforms all other methods in chitchat quality, recording the highest win rate at 40.86\% against \textsc{GPT-4o} in pairwise evaluation. 
It also demonstrates the most \textit{proactive} behavior, with over one successful mode switch and recovery per dialogue---i.e., 1.322 switch attempts with 1.300 successes, and 0.977 recovery attempts with 0.856 successes on average.
We further analyze its qualitative strengths in \S\ref{subsec:qualitative_analysis}.




In contrast, ICL-based models—both zero-shot and few-shot—consistently underperform across most metrics, highlighting the limitations of prompt-only adaptation in handling complex dialogue transitions. 
Although ICL-FS makes the most switch and recovery attempts, its low success rate suggests a tendency to excessively initiate transitions without sufficient contextual grounding.




In summary, experimental results confirm that DPO---a preference-tuned extension of SFT---effectively replaces modular pipelines with a single robust model that jointly handles intent prediction, dialogue flow control, and chitchat generation.
Appendix~\ref{sec:appendix_models} further shows that these trends hold consistently across different underlying models.
\begin{figure}[t]
    \centering
    \includegraphics[width=0.95\linewidth]{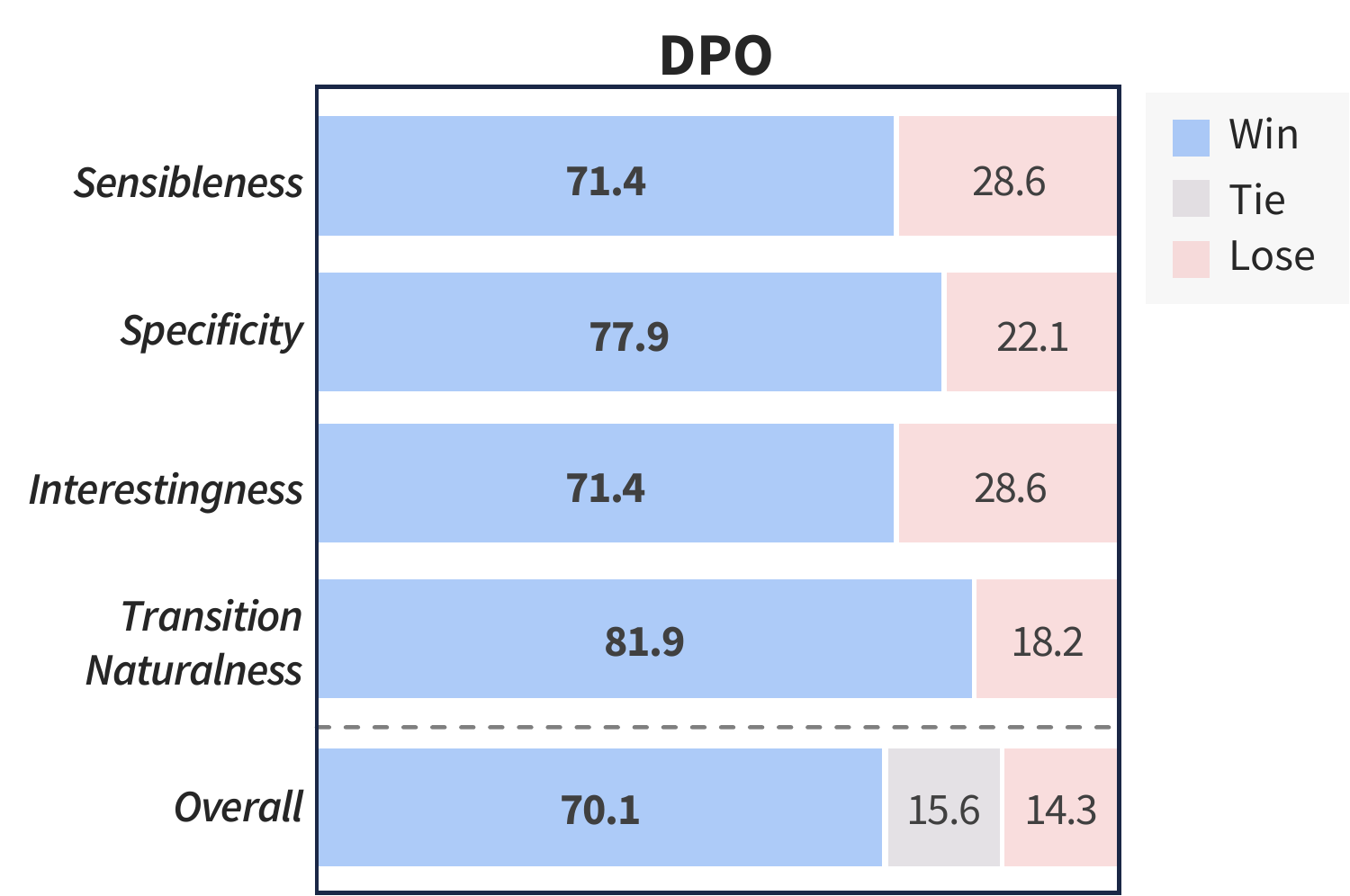}
    \caption{
    Human preference evaluation of the DPO model trained on \tact\ TACT against \textsc{GPT-4o} (few-shot), with over 70\% overall wins across 77 dialogues.
    }
    \label{fig:winrate_human}
\end{figure}

\subsection{Preference-Based Evaluation}
\label{subsec:qualitative_analysis}


Based on the evaluation criteria in \S\ref{sec:chitchat-metrics}, 
we conduct a preference-based analysis of model responses in integrated-mode dialogues, using two types of evaluators: an LLM judge (\textsc{Gemini-2.5-Pro}) and human annotators (10 evaluators, including both NLP practitioners and general users). 
The interface provided to human annotators, which includes evaluation instructions, is shown in Appendix~\ref{sec:appendix_metrics}.
The LLM judge compares responses from each tested model and \textsc{GPT-4o} (few-shot) over the entire test set, producing win/tie/lose outcomes for each criterion (Figure~\ref{fig:generated_winrate}). 
In parallel, the human study was conducted on a 77-sample subset of the test set, where annotators evaluated responses without a tie option, 
providing an additional perspective aligned with real user interactions (Figure~\ref{fig:winrate_human}).\footnote{No ties per criterion, but possible overall (2–2 split).}


DPO outperforms \textsc{GPT-4o} across all four qualitative criteria, with strong gains in \textit{Interestingness} and \textit{Transition Naturalness}. 
These results demonstrate that preference tuning enables models to internalize soft conversational qualities such as engagement and flow continuity. 
In \textit{Interestingness}, DPO more than doubles the win rate, indicating improved expressiveness and user engagement. 
For \textit{Transition Naturalness}, DPO surpasses \textsc{GPT-4o} for the first time with a win rate of 33.7\%, compared to SFT’s 11.3\%.


Human judgments make this contrast even clearer: 
While the LLM judge recorded only around 40\% win rates with many ties, human annotators showed over 70\% wins with sharply reduced ties.\footnote{This is partly due to the no-tie setup, but the significant margins indicate that humans clearly recognize DPO’s merits.}
The advantage holds across all four qualitative criteria, demonstrating that DPO’s gains are overwhelmingly clear to human users.


In addition, case studies with real output examples in Appendix \ref{sec:appendix_transition_samples} show that DPO responses more frequently exhibit contextually anchored reactions\textemdash such as callbacks and affective tone\textemdash and generate smoother transitions within dialogues. 

\begin{table}[t]
    \small
    \centering
    \begin{tabular}{ccaa}
        \toprule
        \textbf{Method} & \textbf{Success} & \textbf{intent \cmark} & \textbf{intent \xmark} \\
        \midrule
        ICL-FS & 0.652 & 33.89 & 66.11 \\
        SFT & 0.856 & 34.58 & 65.42 \\
        SFT-DPO & 0.859 & 34.23 & 65.77 \\
        \bottomrule 
    \end{tabular}
    \caption{
    Recovery Success analysis at mode- and intent-levels. \textbf{intent \cmark} shows the proportion of recoveries that return to the previous intent, and \textbf{intent \xmark} reflects the proportion that initiate a new one.
    }
    \label{tab:recover_ablation}
\end{table}

\begin{table*}
    \small
    \centering
    \resizebox{\textwidth}{!}{%
    \begin{tabular}{laaccaaccaa}
        \toprule
        {} & \multicolumn{6}{>{\cellcolor{blue!10}}c}{\textbf{TOD}} & \multicolumn{4}{>{\cellcolor{yellow!30}}c}{\textbf{Flow}}  \\
        \arrayrulecolor{gray!70} \cmidrule(){2-11} \arrayrulecolor{black}
        \textbf{Flow type}& \multicolumn{2}{a}{\textbf{Mode Selection}} & \multicolumn{2}{c}{\textbf{Intent Detection}} & \multicolumn{2}{a}{\textbf{Joint Accuracy}} &\multicolumn{2}{c}{\textbf{Switch}} & \multicolumn{2}{a}{\textbf{Recovery}}\\
        (\# Dialogues) & \textbf{Acc.} & \textbf{F1} & \textbf{Acc.}/turn & \textbf{Acc.}/dialogue & \textbf{Acc.}/turn & \textbf{Acc.}/dialogue & \textbf{Attempt} & \textbf{Success} & \textbf{Attempt} & \textbf{Success} \\
        \midrule
        \multicolumn{11}{c}{TACT$_\text{MutliWOZ}$ (680)}\\
        \midrule
        TCT (533) & 99.35 & 99.04 & 96.64 & 82.18 & 97.18 & 81.24 & 1.390 & 1.372 & 1.006 & 0.977\\
        CTC (38) & 97.06 & 96.51 & 96.30 & 92.11 & 97.06 & 81.58 & 0.421 & 0.342 & 0.313 & 0.000 \\
        TC (74) & 95.35 & 97.99 & 95.35 & 74.32 & 95.18 & 64.86 & 1.419 & 1.378 & 1.000 & 0.069 \\
        TCTC (27) & 97.35 & 96.94 & 97.39 & 85.19 & 96.46 & 70.37 & 1.222 & 1.185 & 0.958 & 0.958   \\
        Others (7) & 98.36 & 98.29 & 97.22 & 87.50 & 96.72 & 75.00 & 0.500 & 0.500 & 1.000 & 1.000\\
        \midrule
        \multicolumn{11}{c}{TACT$_\text{SLURP}$ (1,790)}\\
        \midrule
        TCT (618) & 99.41 & 99.12 & 94.18 & 72.98 & 95.28 & 72.01 & 0.974 & 0.964 & 1.019 & 0.985 \\
        CTC (907) & 97.61 & 97.00 & 91.69 & 88.20 & 96.10 & 79.82 & 1.756 & 0.821 & 1.041 & 0.037 \\
        TC (60) & 91.19 & 91.01 & 76.68 & 31.67 & 85.32 & 30.00 & 1.450 & 0.550 & 1.304 & 0.071 \\
        TCTC (174) & 96.06 & 96.03 & 91.17 & 71.26 & 92.80 & 60.34 & 1.793 & 1.167 & 1.616 & 0.169 \\
        Others (31) & 85.13 & 84.87 & 82.28 & 45.16 & 84.39 & 16.13 & 1.645 & 0.677 & 1.393 & 0.107 \\
        \bottomrule
    \end{tabular}%
    }
    \caption{
    Performance variation of the SFT agent across dialogue flow types. 
    All extended variants such as TTCT, TCTT, and similar patterns are consolidated into representative flow categories. 
    (e.g., TTCT $\rightarrow$ TCT). 
    }
    \label{tab:flow_specific_results}
\end{table*}

\subsection{Transition-Focused Analysis}



\paragraph{Recovery success analysis}
We further break down the Recovery Success metric (\S \ref{sec:metrics}) by examining whether the resumed task continues the same intent as before the transition.
Table \ref{tab:recover_ablation} reports that only about 34\% of successful recoveries return to the previous intent (\textbf{intent \cmark}), while the rest initiate a new intent within the same mode (\textbf{intent \xmark}).
This indicates that in realistic scenarios, successful recovery does not always require returning to the original intent. 
Depending on the dialogue context, initiating a new but relevant intent can be just as appropriate—as long as the interaction resumes smoothly within the correct mode.

\paragraph{Analysis by dialogue flow type}
In Table~\ref{tab:flow_specific_results}, we present an ablation study evaluating the performance of SFT agents across diverse dialogue mode flows. 
We focus on the SFT agent rather than DPO, since DPO primarily improves response quality in chitchat, which is beyond the scope of this analysis.


The agent performs most reliably on the TCT flow across all metrics, demonstrating high accuracy in mode selection and intent prediction, along with strong switch and recovery behavior. 
In contrast, in the CTC flow, the agent shows a notable gap in transition-aware metrics, especially in ${\text{TACT}_{\text{MultiWOZ}}}$.
As shown in Table \ref{tab:transition_distribution} 
of Appendix \ref{sec:transition_distribution}, 
TACT contains user-driven transitions to chitchat more than agent-driven ones, likely because such transitions were perceived as more natural during the data validation process. 
As a result, the agent shows less transition attempts and recoveries, in CTC settings.
Nevertheless, the agent maintains strong performance on the ToD-centric metrics, indicating that it effectively detects user-driven mode transitions.



The TC flow presents another challenge.
Although recovery is infrequent, the agent still exhibits disproportionately high switch attempt rates (1.450 on SLURP) despite low success rates (0.550 on SLURP).
This suggests mode confusion or over-triggering of switches in cases where the need for transition is minimal or poorly signaled.
The model appears to misinterpret certain turns as transition points, indicating imprecise transition judgment when explicit cues for mode shifts are lacking\footnote{
Note that by construction, the TC flow contains no gold-standard recovery events. 
However, since recovery metrics are defined on model predictions at inference, recovery success can still occur when the model first mispredicts a transition and later corrects it. 
This accounts for the non-zero recovery scores in the TC flow.
}.
Interestingly, the TCTC flow---though structurally more complex---yields stable recovery performance. 
This implies that structural complexity alone does not hinder learning and may even support it when the flow clearly supervises transitions.

Overall, this analysis reveals that while TOD performance (e.g., intent detection) remains stable across flows, flow-sensitive behaviors such as switching and recovery are highly dependent on interaction pattern and training coverage. 
Improving generalization thus requires not only stronger flow-aware supervision, but also better handling of ambiguous or underspecified transitions.

\section{Conclusion}
We present \tact\ TACT, a dataset for modeling the integration of task-oriented and chitchat dialogues with natural mode transitions and recoverable structures.
TACT enables training models that can manage complex dialogue flows, including multi-turn mode shifts and returns to prior tasks.

We also demonstrate that preference optimization via DPO significantly enhances both task accuracy and response quality, outperforming strong baselines across both quantitative and qualitative dimensions.  
In particular, DPO achieves notable gains in transition-sensitive metrics and chitchat preference, showing that soft conversational skills such as engagement and transition smoothness can be effectively learned through preference signals.

Future work includes fine-grained analysis of complex transition flows already present in TACT\textemdash such as TCCT and nested switches\textemdash and extending our modeling framework to more effectively handle such cases.  
We also aim to explore real-time flow tracking and preference adaptation techniques for open-ended agents that must manage dialogue continuity in dynamic settings.

\section*{Limitations}
Despite the promising results, our study has several limitations.  
First, although the TACT dataset includes structurally complex flows---such as TCCT and TCTT---we simplify these into broader categories like TCT during evaluation.  
This aggregation may mask flow-specific behaviors and limit fine-grained performance analysis.  

Second, our qualitative evaluation relies on a single LLM-based judge (\textsc{Gemini-2.5-Pro}), which may introduce biases not fully aligned with human preferences. 
To mitigate this, we also conducted human evaluation alongside the model-based one.
Although we additionally conducted human evaluation, reliance on a single LLM judge still raises concerns about potential bias and imperfect alignment with human judgments.

Finally, we rely solely on DPO for preference tuning; it remains unclear whether the observed gains arise from DPO-specific characteristics or reflect broader advantages of preference-based learning.
Future work could explore other preference optimization strategies beyond DPO—such as reward modeling or ranking-based fine-tuning—to assess whether the observed improvements generalize across different supervision formats.

\section*{Acknowledgements}
This work was supported by Hyundai Motor Company and Kia.
This work was supported by Institute of Information \& communications Technology Planning \& Evaluation (IITP) grant funded by the Korea government(MSIT) (No.RS-2020-II201373, Artificial Intelligence Graduate School Program(Hanyang University)).
This work was supported by Institute of Information \& communications Technology Planning \& Evaluation (IITP) under the artificial intelligence semiconductor support program to nurture the best talents (IITP-2025-RS-2023-00253914) grant funded by the Korea government(MSIT).

\section*{Ethics Statement}
This research was conducted with a commitment to ethical standards in AI development. 
The proposed models and datasets were designed to improve dialogue system performance while minimizing potential biases. 
Our dataset construction relies exclusively on widely used, publicly available benchmarks (MultiWOZ and SLURP), which do not contain personally identifiable information (PII). 
The augmented dialogues were synthetically generated from these sources, further minimizing any risk of privacy leakage. 
While the use of benchmark datasets reduces concerns about privacy, we acknowledge that they may still reflect domain- or culture-specific biases. 
Our work aims to enhance conversational AI capabilities in a responsible and inclusive manner, with particular attention to supporting diverse user groups and \mbox{interaction scenarios.}


\bibliography{anthology, references}

@inproceedings{chu2025sft,
    title={{SFT} Memorizes, {RL} Generalizes: A Comparative Study of Foundation Model Post-training},
    author={Tianzhe Chu and Yuexiang Zhai and Jihan Yang and Shengbang Tong and Saining Xie and Dale Schuurmans and Quoc V Le and Sergey Levine and Yi Ma},
    booktitle={Forty-second International Conference on Machine Learning},
    year={2025},
    url={https://openreview.net/forum?id=dYur3yabMj}
}

@inproceedings{rim-etal-2025-chat,
    title = "To Chat or Task: a Multi-turn Dialogue Generation Framework for Task-Oriented Dialogue Systems",
    author = "Rim, Daniel  and
      Cho, Minsoo  and
      Chun, Changwoo  and
      Choo, Jaegul",
    editor = "Rehm, Georg  and
      Li, Yunyao",
    booktitle = "Proceedings of the 63rd Annual Meeting of the Association for Computational Linguistics (Volume 6: Industry Track)",
    month = jul,
    year = "2025",
    address = "Vienna, Austria",
    publisher = "Association for Computational Linguistics",
    url = "https://aclanthology.org/2025.acl-industry.41/",
    doi = "10.18653/v1/2025.acl-industry.41",
    pages = "576--592",
    ISBN = "979-8-89176-288-6"
}

@article{bocklisch2017rasa,
  title={Rasa: Open source language understanding and dialogue management},
  author={Bocklisch, Tom and Faulkner, Joey and Pawlowski, Nick and Nichol, Alan},
  journal={arXiv preprint arXiv:1712.05181},
  year={2017}
}

@inproceedings{liu2021benchmarking,
  title={Benchmarking natural language understanding services for building conversational agents},
  author={Liu, Xingkun and Eshghi, Arash and Swietojanski, Pawel and Rieser, Verena},
  booktitle={Increasing naturalness and flexibility in spoken dialogue interaction: 10th international workshop on spoken dialogue systems},
  pages={165--183},
  year={2021},
  organization={Springer}
}

@article{thoppilan2022lamda,
  title={Lamda: Language models for dialog applications},
  author={Thoppilan, Romal and De Freitas, Daniel and Hall, Jamie and Shazeer, Noam and Kulshreshtha, Apoorv and Cheng, Heng-Tze and Jin, Alicia and Bos, Taylor and Baker, Leslie and Du, Yu and others},
  journal={arXiv preprint arXiv:2201.08239},
  year={2022}
}

@article{Young_Xing_Pandelea_Ni_Cambria_2022, title={Fusing Task-Oriented and Open-Domain Dialogues in Conversational Agents}, volume={36}, url={https://ojs.aaai.org/index.php/AAAI/article/view/21416}, DOI={10.1609/aaai.v36i10.21416}, abstractNote={The goal of building intelligent dialogue systems has largely been separately pursued under two paradigms: task-oriented dialogue (TOD) systems, which perform task-specific functions, and open-domain dialogue (ODD) systems, which focus on non-goal-oriented chitchat. The two dialogue modes can potentially be intertwined together seamlessly in the same conversation, as easily done by a friendly human assistant. Such ability is desirable in conversational agents, as the integration makes them more accessible and useful. Our paper addresses this problem of fusing TODs and ODDs in multi-turn dialogues. Based on the popular TOD dataset MultiWOZ, we build a new dataset FusedChat, by rewriting the existing TOD turns and adding new ODD turns. This procedure constructs conversation sessions containing exchanges from both dialogue modes. It features inter-mode contextual dependency, i.e., the dialogue turns from the two modes depend on each other. Rich dependency patterns such as co-reference and ellipsis are included. The new dataset, with 60k new human-written ODD turns and 5k re-written TOD turns, offers a benchmark to test a dialogue model’s ability to perform inter-mode conversations. This is a more challenging task since the model has to determine the appropriate dialogue mode and generate the response based on the inter-mode context. However, such models would better mimic human-level conversation capabilities. We evaluate two baseline models on this task, including the classification-based two-stage models and the two-in-one fused models. We publicly release FusedChat and the baselines to propel future work on inter-mode dialogue systems.}, number={10}, journal={Proceedings of the AAAI Conference on Artificial Intelligence}, author={Young, Tom and Xing, Frank and Pandelea, Vlad and Ni, Jinjie and Cambria, Erik}, year={2022}, month={Jun.}, pages={11622-11629} }

@article{hosseini2020simple,
  title={A simple language model for task-oriented dialogue},
  author={Hosseini-Asl, Ehsan and McCann, Bryan and Wu, Chien-Sheng and Yavuz, Semih and Socher, Richard},
  journal={Advances in Neural Information Processing Systems},
  volume={33},
  pages={20179--20191},
  year={2020}
}

@inproceedings{kwon2023efficient,
  title={Efficient memory management for large language model serving with pagedattention},
  author={Kwon, Woosuk and Li, Zhuohan and Zhuang, Siyuan and Sheng, Ying and Zheng, Lianmin and Yu, Cody Hao and Gonzalez, Joseph and Zhang, Hao and Stoica, Ion},
  booktitle={Proceedings of the 29th Symposium on Operating Systems Principles},
  pages={611--626},
  year={2023}
}

@article{yi2024survey,
  title={A Survey on Recent Advances in LLM-Based Multi-turn Dialogue Systems},
  author={Yi, Zihao and Ouyang, Jiarui and Liu, Yuwen and Liao, Tianhao and Xu, Zhe and Shen, Ying},
  journal={arXiv preprint arXiv:2402.18013},
  year={2024}
}

@article{xu2024large,
  title={Large language models are active critics in nlg evaluation},
  author={Xu, Shuying and Hu, Junjie and Jiang, Ming},
  journal={arXiv preprint arXiv:2410.10724},
  year={2024}
}

@article{acikgoz2025desideratum,
  title={A Desideratum for Conversational Agents: Capabilities, Challenges, and Future Directions},
  author={Acikgoz, Emre Can and Qian, Cheng and Wang, Hongru and Dongre, Vardhan and Chen, Xiusi and Ji, Heng and Hakkani-T{\"u}r, Dilek and Tur, Gokhan},
  journal={arXiv preprint arXiv:2504.16939},
  year={2025}
}

@inproceedings{RastogiZSGK20,
  author       = {Abhinav Rastogi and
                  Xiaoxue Zang and
                  Srinivas Sunkara and
                  Raghav Gupta and
                  Pranav Khaitan},
  title        = {Towards Scalable Multi-Domain Conversational Agents: The Schema-Guided
                  Dialogue Dataset},
  booktitle    = {The Thirty-Fourth {AAAI} Conference on Artificial Intelligence, {AAAI}
                  2020, The Thirty-Second Innovative Applications of Artificial Intelligence
                  Conference, {IAAI} 2020, The Tenth {AAAI} Symposium on Educational
                  Advances in Artificial Intelligence, {EAAI} 2020, New York, NY, USA,
                  February 7-12, 2020},
  pages        = {8689--8696},
  publisher    = {{AAAI} Press},
  year         = {2020},
  url          = {https://doi.org/10.1609/aaai.v34i05.6394},
  doi          = {10.1609/AAAI.V34I05.6394},
  bibsource    = {dblp computer science bibliography, https://dblp.org}
}

@article{acikgoz2025td,
  title={TD-EVAL: Revisiting Task-Oriented Dialogue Evaluation by Combining Turn-Level Precision with Dialogue-Level Comparisons},
  author={Acikgoz, Emre Can and Guo, Carl and Dey, Suvodip and Datta, Akul and Kim, Takyoung and Tur, Gokhan and Hakkani-T{\"u}r, Dilek},
  journal={arXiv preprint arXiv:2504.19982},
  year={2025}
}

@article{rafailov2023direct,
  title={Direct preference optimization: Your language model is secretly a reward model},
  author={Rafailov, Rafael and Sharma, Archit and Mitchell, Eric and Manning, Christopher D and Ermon, Stefano and Finn, Chelsea},
  journal={Advances in Neural Information Processing Systems},
  volume={36},
  pages={53728--53741},
  year={2023}
}

@article{sun2020adding,
  title={Adding chit-chat to enhance task-oriented dialogues},
  author={Sun, Kai and Moon, Seungwhan and Crook, Paul and Roller, Stephen and Silvert, Becka and Liu, Bing and Wang, Zhiguang and Liu, Honglei and Cho, Eunjoon and Cardie, Claire},
  journal={arXiv preprint arXiv:2010.12757},
  year={2020}
}

@inproceedings{stricker2024few,
  title={A Few-shot Approach to Task-oriented Dialogue Enhanced with Chitchat},
  author={Stricker, Armand and Paroubek, Patrick},
  booktitle={Proceedings of the 25th Annual Meeting of the Special Interest Group on Discourse and Dialogue},
  pages={590--602},
  year={2024}
}

@article{li2024large,
  title={Large language models as zero-shot dialogue state tracker through function calling},
  author={Li, Zekun and Chen, Zhiyu Zoey and Ross, Mike and Huber, Patrick and Moon, Seungwhan and Lin, Zhaojiang and Dong, Xin Luna and Sagar, Adithya and Yan, Xifeng and Crook, Paul A},
  journal={arXiv preprint arXiv:2402.10466},
  year={2024}
}

@article{li2025preference,
  title={Preference Leakage: A Contamination Problem in LLM-as-a-judge},
  author={Li, Dawei and Sun, Renliang and Huang, Yue and Zhong, Ming and Jiang, Bohan and Han, Jiawei and Zhang, Xiangliang and Wang, Wei and Liu, Huan},
  journal={arXiv preprint arXiv:2502.01534},
  year={2025}
}

@article{wataoka2024self,
  title={Self-preference bias in llm-as-a-judge},
  author={Wataoka, Koki and Takahashi, Tsubasa and Ri, Ryokan},
  journal={arXiv preprint arXiv:2410.21819},
  year={2024}
}

@article{dubois2024length,
  title={Length-controlled alpacaeval: A simple way to debias automatic evaluators},
  author={Dubois, Yann and Galambosi, Bal{\'a}zs and Liang, Percy and Hashimoto, Tatsunori B},
  journal={arXiv preprint arXiv:2404.04475},
  year={2024}
}

@inproceedings{hu2024explaining,
  title={Explaining length bias in llm-based preference evaluations},
  author={Hu, Zhengyu and Song, Linxin and Zhang, Jieyu and Xiao, Zheyuan and Chen, Zhengyu and Xiong, Hui},
  booktitle={ICLR 2025 Workshop on Navigating and Addressing Data Problems for Foundation Models},
  year={2024}
}

@article{hurst2024gpt,
  title={Gpt-4o system card},
  author={Hurst, Aaron and Lerer, Adam and Goucher, Adam P and Perelman, Adam and Ramesh, Aditya and Clark, Aidan and Ostrow, AJ and Welihinda, Akila and Hayes, Alan and Radford, Alec and others},
  journal={arXiv preprint arXiv:2410.21276},
  year={2024}
}

@misc{gemini2023,
  author    = {Google DeepMind},
  title     = {Gemini: Multimodal and Language Model},
  year      = {2023},
  note      = {Accessed: 2023-09-05},
  url       = {https://www.deepmind.com/research/gemini}
}

@article{grattafiori2024llama,
  title={The llama 3 herd of models},
  author={Grattafiori, Aaron and Dubey, Abhimanyu and Jauhri, Abhinav and Pandey, Abhinav and Kadian, Abhishek and Al-Dahle, Ahmad and Letman, Aiesha and Mathur, Akhil and Schelten, Alan and Vaughan, Alex and others},
  journal={arXiv preprint arXiv:2407.21783},
  year={2024}
}

@misc{deepspeed2021,
  author    = {DeepSpeed},
  title     = {DeepSpeed ZeRO-3 Offload},
  year      = {2021},
  note      = {Accessed: 2021-03-07},
  url       = {https://www.deepspeed.ai/2021/03/07/zero3-offload.html}
}

@inproceedings{papineni2002bleu,
  title={Bleu: a method for automatic evaluation of machine translation},
  author={Papineni, Kishore and Roukos, Salim and Ward, Todd and Zhu, Wei-Jing},
  booktitle={Proceedings of the 40th annual meeting of the Association for Computational Linguistics},
  pages={311--318},
  year={2002}
}

\vfill\break
\appendix






\onecolumn
\appendix
\part*{Appendix}


\section*{Appendix Contents}

\begin{itemize}
    \item Appendix~\ref{sec:appendix_tact_construction} \tact\ TACT Construction 
    \begin{itemize}
        \item Appendix~\ref{sec:appendix_startegy_case_study} Baseline vs. Proposed Strategies        
        \item Appendix~\ref{sec:flow_construction} TCT/CTC Dialogue Construction
        \item Appendix~\ref{sec:transition_distribution} Dialogue Statistics
    \end{itemize}
    \item Appendix~\ref{sec:appendix_tact_validation} \tact\ TACT Validation
    \begin{itemize}
        \item Appendix~\ref{sec:appendix_val_proc} Validation Procedure
        \item Appendix~\ref{sec:appendix_val_results} Validation Results
    \end{itemize}
    \item Appendix~\ref{sec:appendix_modeling} Training Details
    \begin{itemize}
        \item Appendix~\ref{sec:appendix_icl_prompts} Prompts for In-Context Learning
        \item Appendix~\ref{sec:appendix_dpo_samples} DPO Response Pair Examples
    \end{itemize}
    \item Appendix~\ref{sec:appendix_experiment} Experimental Setup
    \begin{itemize}
        \item Appendix~\ref{sec:appendix_env} Environments
        \item Appendix~\ref{sec:appendix_metrics} Evaluation Metrics
    \end{itemize}
    \item Appendix~\ref{sec:appendix_discussion} Supplementary Results
    \begin{itemize}
        \item Appendix~\ref{sec:appendix_preliminary_eval} Preliminary Results
        \item Appendix~\ref{sec:appendix_models} Backbone Comparison
        \item Appendix~\ref{sec:appendix_slurp} $\text{TACT}_\text{SLURP}$ Analysis
        \item Appendix~\ref{sec:appendix_transition_samples} Examples of Transition Handling
    \end{itemize}
\end{itemize}

\clearpage
\section{\tact\ TACT Construction} \label{sec:appendix_tact_construction}


\subsection{Baseline vs. Proposed Strategies}
\label{sec:appendix_startegy_case_study}
The dialogues in Table~\ref{tab:appendix_strategy_comparison} embody three \textit{transition-oriented} generation strategies that overcome the limitations of our two source corpora. 
\textbf{MultiWOZ 2.2} already contains multi-turn TOD segments, but they rarely mix with chitchat. 
In contrast, \textbf{SLURP} consists only of \textit{single-turn} intent utterances, which are too sparse to train a model that must handle multi-turn TOD or mode shifts.
\\

\textbf{Strategy 1. Utterance Expansion (Short TOD)}---
Starting from a single SLURP intent, we add one or two open-ended questions to elicit a concrete request, yielding a concise short TOD that serves as a building block for later CTC construction.

\textbf{Strategy 2. Intent-Scenario Expansion (Long TOD)}--- We chain semantically related intents (e.g., ``play~podcast'' $\rightarrow$ ``play~music'') to extend the short TOD into a \textbf{long, coherent} TOD, enabling more realistic multi-intent conversations and stronger modeling of sub-tasks.

\textbf{Strategy 3. Chitchat Transitions (TCT/CTC)}---
With a TOD core in place (from Strategy 1 or 2), we prepend or append brief chitchat turns to construct controlled mode-shift patterns. 
In the \textbf{TCT} setting, a chitchat detour is inserted into a long TOD (from \textbf{Strategy 2}), whereas in the \textbf{CTC} setting, a short TOD (from \textbf{Strategy 1}) is wrapped with small-talk openings and closings.
\\

This three-step process (i) expands SLURP’s single-turn data into rich multi-turn TOD, and (ii) grafts chitchat segments onto TOD from either corpus, producing diverse TCT and CTC flows. 
Together, these strategies yield \emph{transition-optimized} training data that raw MultiWOZ 2.2 and SLURP alone cannot provide.


\begin{table}[H]
    \centering
    \small
    \renewcommand{\arraystretch}{1}
    \rowcolors{2}{gray!10}{white}
    \resizebox{\linewidth}{!}{
    \begin{tabular}{m{0.18\linewidth} m{0.82\linewidth}}
        \toprule
        \textbf{Input (\texttt{Intent})} & \makecell[l]{\texttt{please play npr podcast} (\texttt{play\_podcasts})} \\
        \midrule
        \rowcolor{gray!25}
        \multicolumn{2}{l}{\textbf{Baseline Strategy (No Transition Optimization)}} \\
        \midrule
        \textbf{Baseline Response} & 
        \begin{minipage}[t]{\linewidth}
            \textbf{User}: I've been really into podcasts lately, especially news ones. \\
            \textbf{Agent}: That's great! Are you looking for something specific today, or do you have a favorite show in mind? \\
            \textbf{User}: Please play NPR podcast. \\
            \textbf{Agent}: Sure! I can play the latest episode of NPR for you. Would you prefer the daily news, or is there a particular topic you're interested in? \\
            \textbf{User}: I'm in the mood for the daily news today. \\
            \textbf{Agent}: Got it! Playing the latest NPR daily news episode now.
        \end{minipage} \\
        \midrule
        \rowcolor{gray!25}
        \multicolumn{2}{l}{\textbf{Proposed Strategies with Transition Optimization}} \\
        \midrule
        \textbf{Strategy 1. Utterance Expansion} & 
        \begin{minipage}[t]{\linewidth}
            \textbf{User}: I've been wanting to listen to something interesting. Can you help me find a podcast? \\
            \textbf{Agent}: Absolutely! What kind of topics are you interested in? \\
            \textbf{User}: Please play NPR podcast.
        \end{minipage} \\
        \midrule
        \textbf{Strategy 2. Intent Scenario Expansion} & 
        \begin{minipage}[t]{\linewidth}
            \textit{(Continuing from Strategy 1)} \\
            \textbf{Agent}: Sure! Playing the latest episode of the NPR podcast now. Is there anything else you’d like to listen to after this? \\
            \textbf{User}: Yes, I'd like to listen to some music after the podcast. \\
            \textbf{Agent}: Great! What genre of music do you prefer? \\
            \textbf{User}: I love to hear some jazz. \\
            \textbf{Agent}: Playing some smooth jazz for you after the NPR podcast. Would you like to set a specific duration for the music? \\
            \textbf{User}: Just play it for an hour, please.
        \end{minipage} \\
        \midrule
        \textbf{Strategy 3. Chitchat Transitions} & 
        \begin{minipage}[t]{\linewidth}
            \textit{(Continuing from Strategy 1)} \\
            \textbf{Agent}: Sure! Playing the latest episode of the NPR podcast now. \\
            \textbf{User}: I really enjoy jazz; it's so relaxing. Do you think it's a good genre for unwinding? \\
            \textbf{Agent}: Jazz is known for its soothing melodies and improvisation, making it perfect for relaxation. \\
            \textbf{User}: I also love how it can set a nice ambiance for a cozy evening at home. \\
            \textbf{Agent}: It sounds like you appreciate a calming atmosphere! Would you like to set a specific duration for the jazz music after the podcast? \\
            \textbf{User}: Just play it for an hour, please.
        \end{minipage} \\
        \bottomrule
    \end{tabular}
    }
    \caption{
    Comparison of baseline and transition-optimized strategies under a shared user intent (\texttt{play\_podcast}). 
    The baseline yields a rigid TOD exchange, while the proposed strategies demonstrate more natural dialogue through 
    (i) utterance expansion, (ii) intent-scenario expansion, and (iii) chitchat transitions (TCT/CTC).
    }
    \label{tab:appendix_strategy_comparison}
\end{table}

\raggedbottom

\subsection{TCT/CTC Dialogue Construction} \label{sec:flow_construction}

We construct two types of transition patterns, \textbf{TCT} and \textbf{CTC}, and both patterns are realized on two corpora, SLURP and MultiWOZ2.2, with dataset-specific procedures.

\begin{figure}[H]
    \centering
    \includegraphics[scale=0.95]{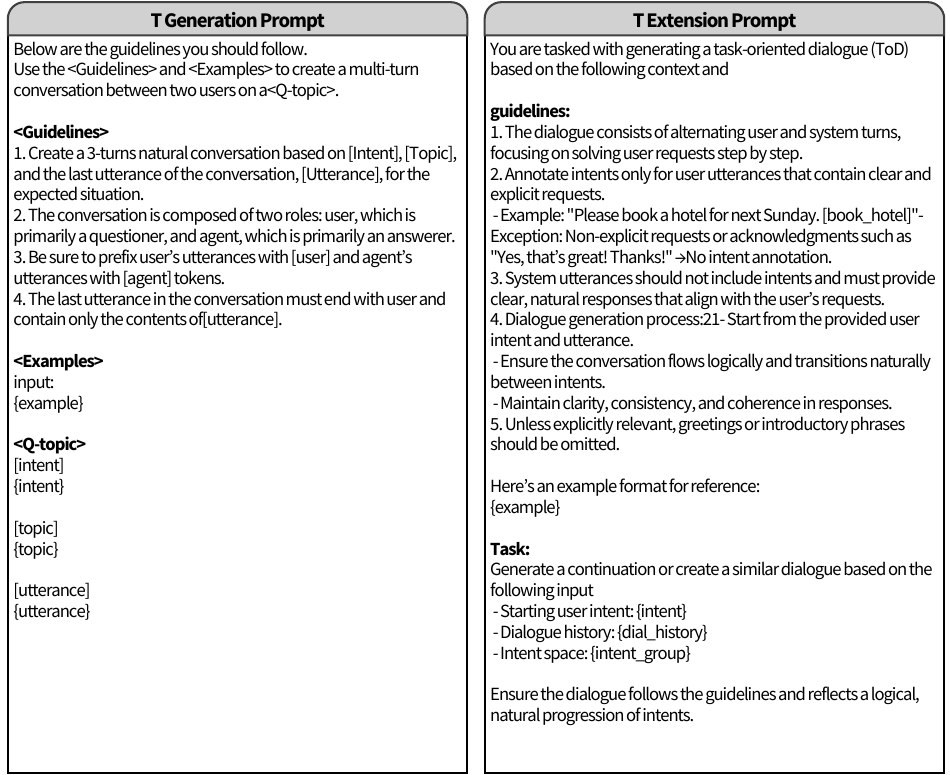}
    \caption{
    Prompts for SLURP-based TOD construction. 
    The generation prompt converts single-turn SLURP utterances into short TODs, and the extension prompt expands these short TODs into longer multi-turn dialogues by following an intent flow. 
    \texttt{\{intent\_group\}} denotes the set of candidate intents from which the continuation must be chosen (see Table~\ref{tab:intent_flow_full} for the full intent flow specification).
     }
    \label{fig:prompt_slurp_t_extension}
\end{figure}

\begin{figure}[H]
    \centering
    \includegraphics[scale=1]{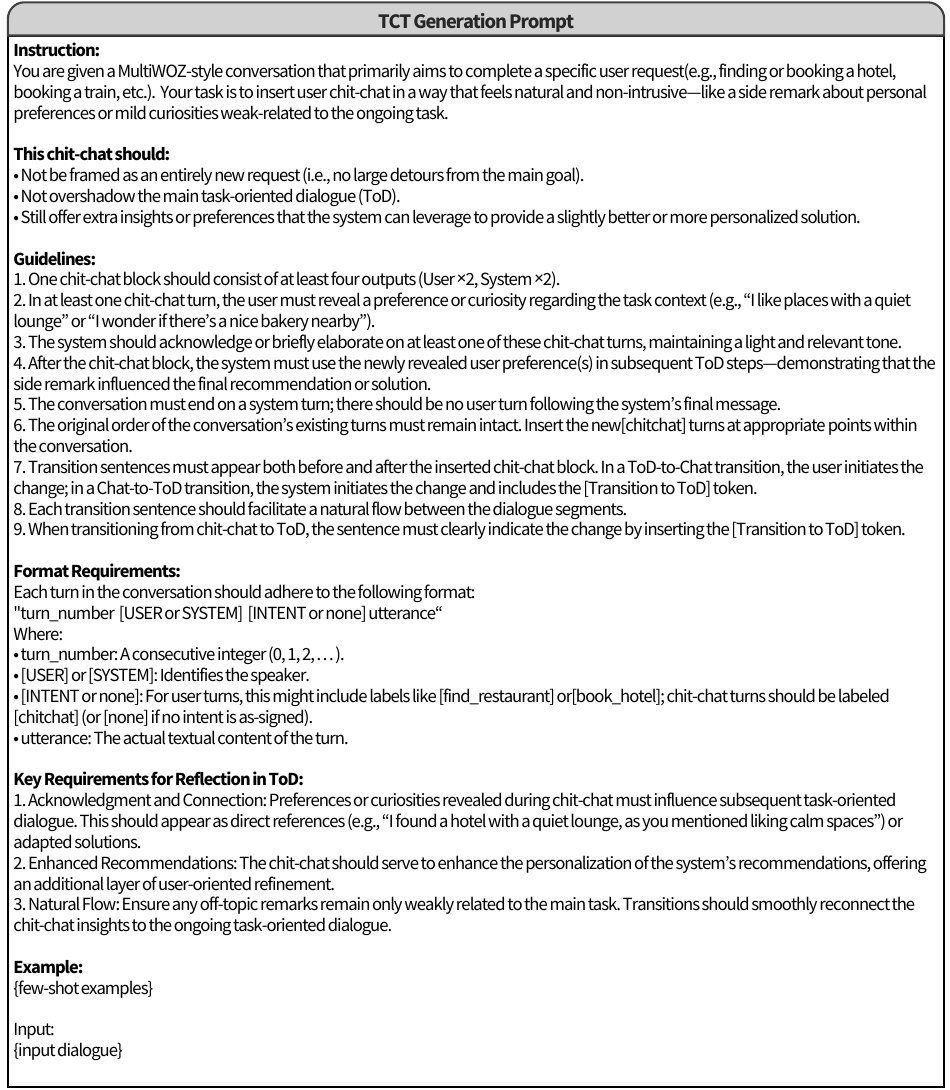}
    \caption{Prompts for TCT dialogue construction. 
    Given a TOD core (from MultiWOZ long segments or SLURP-extended flows), 
    a brief chitchat utterance is inserted in the middle to form a TOD–Chitchat–TOD transition.
    }
    \label{fig:prompt_tct}
\end{figure}

\paragraph{TCT (TOD–Chitchat–TOD).} 
\begin{itemize} 
    \item SLURP: Since SLURP consists of single-turn intents, we first chain semantically related intents to form short TODs of about three turns. 
    These are then expanded into longer flows (Table~\ref{tab:intent_flow_full}, Figure~\ref{fig:prompt_slurp_t_extension}), and a chitchat is inserted mid-dialogue to mimic a realistic interruption (Figure~\ref{fig:prompt_tct}).
    \item MultiWOZ2.2: We select long TOD segments (at least four turns) to serve as the core of the dialogue, then insert a brief chitchat in the middle to simulate a natural conversational interruption (Figure~\ref{fig:prompt_tct}).
\end{itemize}

\begin{figure}[H]
    \centering
    \includegraphics[scale=1]{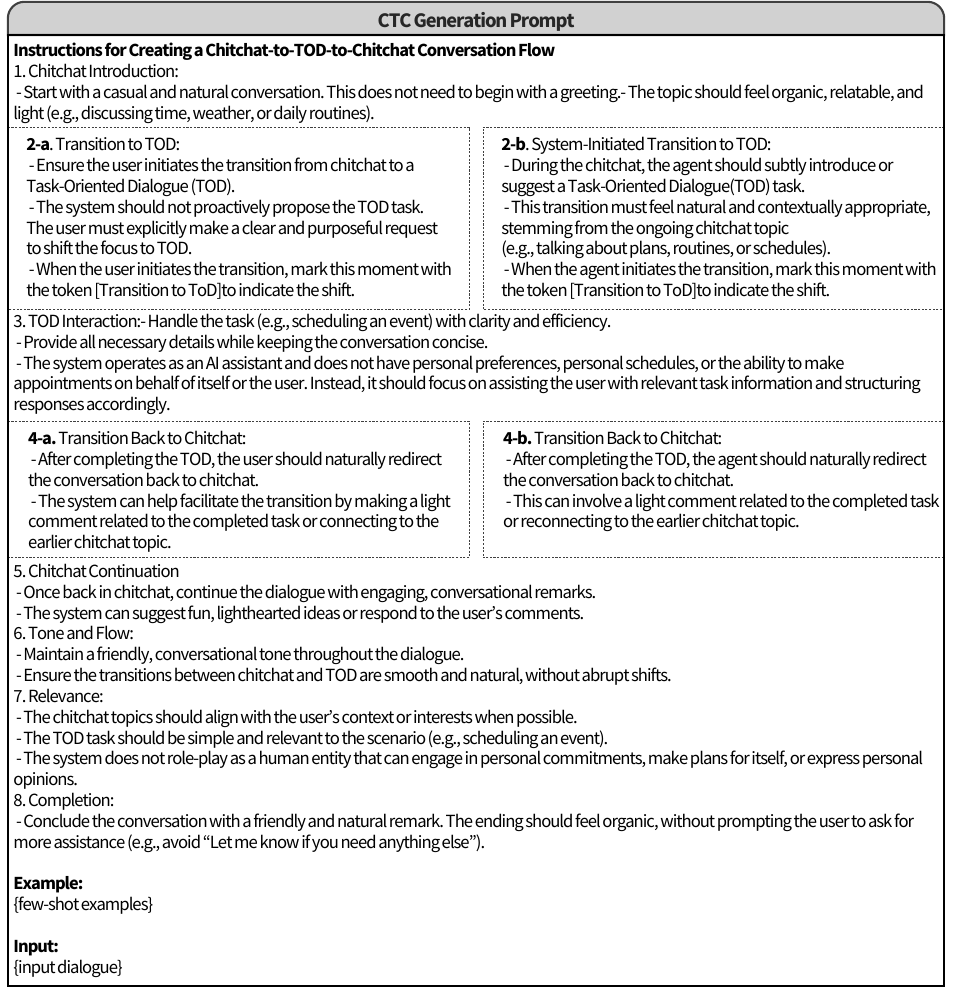}
    \caption{Prompt for CTC dialogue construction.
    A short TOD core (up to 3 turns) is wrapped with chitchat utterances at both the beginning and the end, forming a Chitchat–TOD–Chitchat flow. 
    Two variants are illustrated: \textit{User-Initiated} (2-a, 4-a), where the user opens and closes the task, 
    and \textit{Agent-Initiated} (2-b, 4-b), where the system guides the transitions.
    }
    \label{fig:prompt_ctc}
\end{figure}

\paragraph{CTC (Chitchat–TOD–Chitchat).} 
\begin{itemize}
    \item SLURP: We follow the same chaining procedure as in the TCT construction to build short TODs, and then add chitchat turns before and after the TOD to simulate casual conversation framing.
    \item MultiWOZ2.2: We select short TOD segments with fewer than four turns and wrap them with brief chitchat at both the beginning and the end, embedding the task in open-domain dialogue (Figure~\ref{fig:prompt_ctc}). 
\end{itemize}
This construction strategy produces varied dialogue sequences combining open-domain and task-oriented behaviors, thereby supporting robust training and evaluation of transition-heavy interactions.

\raggedbottom
\begin{table*}[t]
    \small
    \centering
    \begin{tabular}{ll} 
        \toprule
        \rowcolor{gray!10}\textbf{Intent} & \textbf{Intent Flow} \\
        \midrule
        set\_event & remove\_event, check\_weather, send\_email, play\_music, recommend\_events \\
        check\_calendar & remove\_event, set\_event, send\_email, check\_weather, check\_lists \\
        remove\_event & set\_event, check\_calendar, send\_email, play\_music, recommend\_events \\
        increase\_volume & decrease\_volume, mute\_volume, change\_volume, play\_music, play\_game \\
        mute\_volume & increase\_volume, decrease\_volume, change\_volume, play\_music, play\_game \\
        decrease\_volume & increase\_volume, change\_volume, mute\_volume, play\_music, play\_game \\
        change\_volume & increase\_volume, decrease\_volume, mute\_volume, play\_music, play\_game \\
        increase\_light & dim\_light, turnon\_light, change\_light, turnoff\_light, make\_coffee \\
        make\_coffee & start\_cleaner, turnon\_wemo, turnoff\_wemo, turnon\_light, book\_taxi \\
        start\_cleaner & turnoff\_wemo, make\_coffee, turnon\_light, dim\_light, turnon\_wemo \\
        turnon\_wemo & turnoff\_wemo, make\_coffee, start\_cleaner, turnon\_light, change\_light \\
        change\_light & dim\_light, turnoff\_light, turnon\_light, make\_coffee, start\_cleaner \\
        turnoff\_light & turnon\_light, dim\_light, change\_light, make\_coffee, start\_cleaner \\
        turnoff\_wemo & turnon\_wemo, make\_coffee, start\_cleaner, turnon\_light, change\_light \\
        turnon\_light & dim\_light, turnoff\_light, change\_light, make\_coffee, start\_cleaner \\
        dim\_light & increase\_light, turnon\_light, change\_light, make\_coffee, start\_cleaner \\
        check\_weather & set\_event, recommend\_events, book\_taxi, play\_music, send\_email \\
        remove\_list & createoradd\_list, check\_lists, send\_email, check\_email, find\_recipe \\
        createoradd\_list & check\_lists, remove\_list, send\_email, check\_email, find\_recipe \\
        check\_lists & createoradd\_list, remove\_list, send\_email, check\_email, find\_recipe \\
        check\_email & send\_email, check\_contact, add\_contact, set\_alarm, check\_alarm \\
        add\_contact & check\_contact, send\_email, set\_event, check\_email, set\_alarm \\
        send\_email & check\_email, add\_contact, set\_alarm, check\_contact, set\_event \\
        check\_contact & send\_email, add\_contact, set\_event, check\_email, set\_alarm \\
        set\_alarm & check\_alarm, remove\_alarm, send\_email, check\_email, set\_event \\
        check\_alarm & set\_alarm, remove\_alarm, send\_email, check\_email, set\_event \\
        remove\_alarm & set\_alarm, check\_alarm, send\_email, check\_email, set\_event \\
        play\_music & play\_podcast, play\_audiobook, play\_radio, play\_game, play\_podcast \\
        play\_podcast & play\_music, play\_audiobook, play\_radio, play\_game, play\_audiobook \\
        play\_audiobook & play\_podcast, play\_music, play\_radio, play\_game, play\_podcast \\
        play\_radio & play\_music, play\_podcast, play\_audiobook, play\_game, play\_music \\
        play\_game & play\_music, play\_podcast, play\_audiobook, play\_radio, play\_podcast \\
        recommend\_events & recommend\_locations, recommend\_movies, check\_weather, play\_music, recommend\_movies \\
        recommend\_locations & recommend\_events, recommend\_movies, check\_weather, play\_music, recommend\_movies \\
        recommend\_movies & recommend\_events, recommend\_locations, check\_weather, play\_music, recommend\_events \\
        check\_social & post\_social, check\_news, play\_music, order\_food \\
        post\_social & check\_social, check\_news, play\_music, order\_food \\
        check\_news & check\_social, post\_social, play\_music, order\_food \\
        adjust\_music & query\_music, play\_music, order\_food \\
        query\_music & adjust\_music, play\_music, order\_food \\
        order\_food & check\_food, book\_taxi, check\_transport, play\_music, check\_transport \\
        check\_food & order\_food, book\_taxi, check\_transport, play\_music, check\_transport \\
        check\_transport & book\_taxi, order\_food, check\_traffic, play\_music, order\_food \\
        book\_taxi & check\_transport, check\_food, check\_traffic, play\_music \\
        book\_ticket & check\_transport, check\_traffic, play\_game, play\_music, check\_transport \\
        check\_traffic & check\_transport, book\_taxi, play\_music, book\_taxi \\
        find\_recipe & ask\_cooking, order\_food, check\_food, play\_music, ask\_cooking \\
        ask\_cooking & find\_recipe, order\_food, check\_food, play\_music, find\_recipe \\
        check\_datetime & convert\_time, set\_event, check\_weather, play\_music, convert\_time \\
        convert\_time & check\_datetime, set\_event, check\_weather, play\_music, check\_weather \\
        \bottomrule
    \end{tabular}
    \caption{
    Intent-to-intent flow mapping for SLURP-based TOD construction.
    To build this schema, we merge semantically overlapping intents (e.g., \texttt{hue\_lightoff} and \texttt{iot\_hue\_lightoff}), 
    rename them into more intuitive verb-style labels, and reduce the set from the original 93 SLURP intents to 50. 
    We also exclude intents from the QA topic, as they are closer to open-domain chitchat and thus not suitable for task-oriented flow design. 
    The resulting schema defines all allowable transitions between intents, which are used to expand single-turn SLURP utterances into coherent multi-turn TODs
    (see Figure~\ref{fig:dialogue_generation}).
    }
    \label{tab:intent_flow_full} 
\end{table*}

\FloatBarrier
\subsection{Dialogue Statistics}
\label{sec:transition_distribution}

Unlike prior datasets that restrict transitions to user-initiated switches or fix the initiator type within each flow, \tact\ TACT incorporates both user- and agent-driven transitions. 
This design expands the range of conversational dynamics and enables the training of agent-driven, proactive mode transitions. 
As summarized in Table~\ref{tab:transition_distribution}, TACT\textsubscript{SLURP} exhibits the most balanced distribution, with agent-initiated switches averaging over 0.7 per dialogue in both TOD$\mapsto$Chitchat (TC) and Chitchat$\mapsto$TOD (CT) directions.

\begin{table}[H]
    \small
    \centering
    \setlength{\tabcolsep}{2pt}
    \begin{tabular}{cccaca}
        \toprule
        \multicolumn{2}{l}{\textbf{Flow type ($\rightarrow$)}} & \multicolumn{2}{c}{\textbf{TC}} & \multicolumn{2}{c}{\textbf{CT}} \\
        \cmidrule(lr){3-4} \cmidrule(lr){5-6}
        \multicolumn{2}{l}{\textbf{Datasets ($\downarrow$)}} & \textbf{User-driven} & \textbf{Agent-driven} & \textbf{User-driven} & \textbf{Agent-driven} \\
        \midrule
        \multicolumn{2}{c}{FusedChat} & 0.55 & 0.00 & 0.43 & 0.00 \\
        \multicolumn{2}{c}{InterfereChat} & 1.00* & 0.00 & 0.00 & 1.00* \\
        \midrule
        \multirow{3}{*}{\textbf{${\text{TACT}_{\text{MultiWOZ}}}$}}
         & TCT & 0.48 & 0.51 & 0.18 & 0.82 \\
         & CTC & 1.00 & 0.00 & 0.70 & 0.30 \\
         & \textbf{Overall} & \textbf{0.59} & \textbf{0.46} & \textbf{0.23} & \textbf{0.66} \\
        \midrule
        \multirow{3}{*}{\textbf{${\text{TACT}_{\text{SLURP}}}$}}
         & TCT & 0.94 & 0.05 & 0.04 & 0.96 \\
         & CTC & 0.92 & 0.08 & 0.54 & 0.45 \\
         & \textbf{Overall} & \textbf{1.00} & \textbf{0.09} & \textbf{0.32} & \textbf{0.65} \\
        \bottomrule
    \end{tabular} 
    \caption{Transition frequencies at mode-shift points. 
    Average number of TC and CT transitions per dialogue, segmented by the initiator of the switch (user or agent). 
    Asterisks (*) indicate that in InterfereChat, all CT shifts are user-driven and all TC shifts are agent-driven by design.
    }
    \label{tab:transition_distribution}
\end{table}

Table \ref{tab:Flow cases across datasets} rovides the counts underlying Figure \ref{fig:flow_histogram}.
In TACT\textsubscript{MultiWOZ}, the most common flow is TCT, which makes up roughly 72\%, whereas in TACT\textsubscript{SLURP}, the most common is CTC at around 54\%.

\begin{table*}[h]
    \small
    \centering
    \resizebox{\textwidth}{!}{%
    \begin{tabular}{ccrrrrrrrrrrrrrra}
        \toprule
        \rowcolor{gray!10} & & C & T & CT & TC & CTC & TCT & CTCT & TCTC & CTCTC & TCTCT & CTCTCT & TCTCTC & CTCTCTC & TCTCTCT & \textbf{Total} \\
        \midrule
        \multirow{4}{*}{\textbf{${\text{TACT}_{\text{MultiWOZ}}}$}} & Train  & 1 & 7 & 1 & 754 & 655 & 4,111 & 8 & 209 & 29 & 28 & 0 & 8 & 0 & 0 & 5,811 \\
         & Dev     & 0 & 0 & 0 & 81 & 23 & 569 & 0 & 29 & 2 & 4 & 0 & 0 & 0 & 0 & 708\\
         & Test    & 0 & 0 & 0 & 74 & 38 & 533 & 1 & 27 & 4 & 2 & 0 & 1 & 0 & 0 & 680\\
        \rowcolor{blue!10} \cellcolor{white} & \textbf{Total}   & 1 & 7 & 1 & 909 & 716 & 5,213 & 9 & 265 & 35 & 34 & 0 & 9 & 0 & 0 & 7,199\\
        \midrule
        \multirow{4}{*}{\textbf{${\text{TACT}_{\text{SLURP}}}$}} & Train      & 0 & 16 & 4 & 190 & 3,826 & 2,298 & 8 & 555 & 18 & 12 & 0 & 4 & 0 & 1 & 6,932\\
         & Dev        & 3 & 1 & 4 & 56 & 601 & 425 & 5 & 109 & 4 & 4 & 0 & 2 & 0 & 0 & 1,214\\
         & Test       & 4 & 3 & 0 & 60 & 907 & 618 & 8 & 174 & 9 & 3 & 0  & 4 & 0 & 0 & 1,790\\
        \rowcolor{blue!10} \cellcolor{white} & \textbf{Total}      & 7 & 20 & 8 & 306 & 5,334 & 3,341 & 21 & 838 & 31 & 19 & 0 & 10 & 0 & 1 & 9,936 \\
        \bottomrule
    \end{tabular}%
    }
    \caption{Distribution of dialogue flow patterns in TACT Datasets.}
    \label{tab:Flow cases across datasets}
\end{table*}

\raggedbottom

\section{\tact\ TACT Validation}
\label{sec:appendix_tact_validation}

\subsection{Validation Procedure}\label{sec:appendix_val_proc}

\begin{figure}[H]
    \centering
    \includegraphics[scale=0.95]{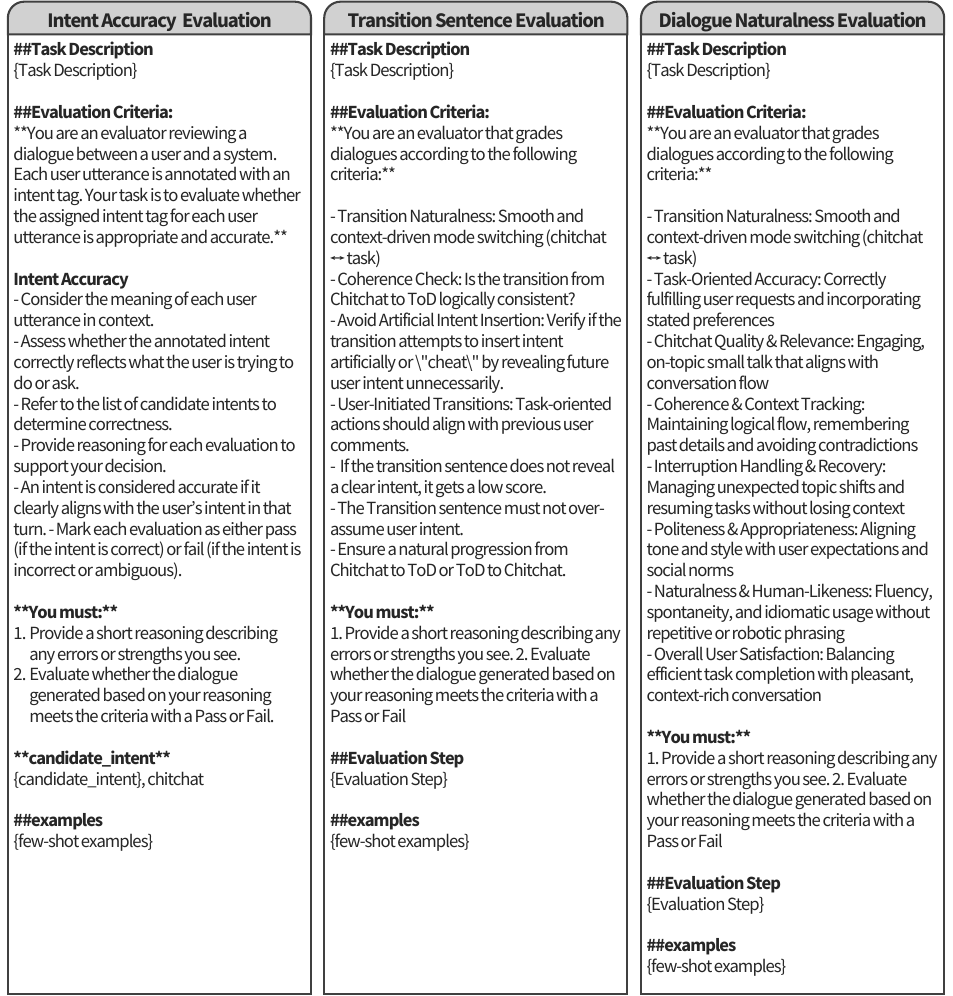}
    \caption{
    Prompts for dialogue validation, covering \textbf{Intent Accuracy}, 
\textbf{Transition Sentence Quality}, and \textbf{Dialogue Naturalness}. 
    All prompts share a common task description, with the latter two including LLM-generated step-by-step instructions. 
    }
\label{fig:appendix_prompt_validation}
\end{figure}

To ensure the quality of our constructed dialogue dataset, 
we performed a three-stage validation focusing on \textbf{Intent Accuracy}, \textbf{Transition Sentence Quality}, and \textbf{Dialogue Naturalness}. 
Each criterion was assessed using structured prompts (Figure~\ref{fig:appendix_prompt_validation}), with the automatic evaluator \textsc{GPT-4o-Mini} providing a \textit{Pass/Fail} judgment and a brief justification. 
For Transition Sentence and Dialogue Naturalness, the prompts further included step-by-step reasoning instructions to enhance judgment consistency. 


Intent Accuracy was measured over overlapping 5-turn windows to capture local misalignments, whereas Transition Sentence and Dialogue Naturalness were assessed at the full-dialogue level. 
A dialogue was retained only if it passed all three criteria. 



\FloatBarrier

\subsection{Validation Results} \label{sec:appendix_val_results}
Filtering was applied independently to each sub-dataset, and the results are reported separately for \textbf{TACT\textsubscript{MultiWOZ}} (Table~\ref{tab:validation_stats_multiwoz}) and \textbf{TACT\textsubscript{SLURP}} (Table~\ref{tab:validation_stats_slurp}).
\begin{table}[H]
    \centering
    \small
    \begin{tabular}{lcca}
        \toprule
        \textbf{Criterion} & \# Validated \cmark & \# Flagged \xmark & \textbf{\# Final Dialogue} (train / dev / test) \\
        \midrule
        Intent Accuracy & 7,840 & 2,303 &  \\
        Dialogue Naturalness & 9,796 & 347 & \textbf{5,811 / 708 / 680}\\
        Transition Sentence Evaluation & 9,531 & 612 & \\
        \bottomrule
    \end{tabular}
    \caption{Validation results for TACT\textsubscript{MultiWOZ}.}
    \label{tab:validation_stats_multiwoz}
\end{table}

\begin{table}[H]
    \centering
    \small
    \begin{tabular}{lcca}
        \toprule
        \textbf{Criterion} & \# Validated \cmark & \# Flagged \xmark & \textbf{\# Final Dialogue} (train / dev / test) \\
        \midrule
        Intent Accuracy & 10,117 & 3,489 &  \\
        Dialogue Naturalness & 13,504 & 102 &  \textbf{6,932 / 1,214 / 1,790} \\
        Transition Sentence Evaluation & 13,476 & 130 & \\
        \bottomrule
    \end{tabular}
    \caption{Validation results for TACT\textsubscript{SLURP}.}
    \label{tab:validation_stats_slurp}
\end{table}

\raggedbottom


\section{Training Methodology Details}\label{sec:appendix_modeling}
\subsection{Prompts for In-Context Learning}
\label{sec:appendix_icl_prompts}
\begin{figure}[H]
    \centering
    \includegraphics[scale=0.95]{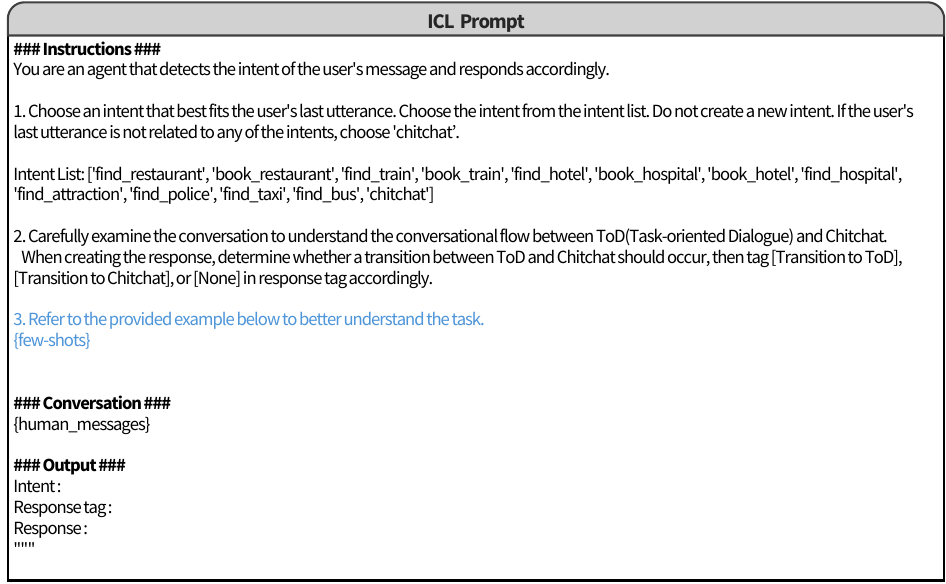}
    \caption{
    ICL prompt template. 
    The prompt specifies intent prediction, transition tagging (\texttt{[Transition to ToD]}, \texttt{[Transition to Chitchat]}, or \texttt{[None]}), and response generation. 
    Illustrated here is the few-shot version, where demonstration turns (\texttt{\{few-shots\}}) are added to the zero-shot template.
    }
    \label{fig:prompt_icl}
\end{figure}

To support ICL evaluation, we adopt both zero-shot and few-shot configurations with manually designed prompts. 
The zero-shot prompt contains only the target input, while the few-shot prompt augments it with a small number of demonstration turns for mode and intent prediction. 
Apart from the presence of these demonstrations, the prompt format remains identical across settings. 
The full template is provided in Figure~\ref{fig:prompt_icl}.

\subsection{DPO Response Pair Examples}
\label{sec:appendix_dpo_samples}

To illustrate how DPO is applied in our framework, we provide a representative example of a dialogue instance used during preference training. For each input dialogue, the model generates two candidate responses—one preferred (chosen) and one less preferred (rejected). These responses are paired to guide the DPO objective, where the model is optimized to assign higher likelihood to the chosen output.

As shown in Table~\ref{tab:appendix_dpo_examples}, both responses are functionally valid and aligned with the user’s request. However, the chosen response is preferred as it provides a more informative and engaging continuation of the dialogue. Compared to the rejected one, it offers richer interaction by extending beyond the immediate question, thereby demonstrating stronger task competence and conversational quality.

\begin{table}[h]
    \centering
    \small
    \renewcommand{\arraystretch}{1}
    \resizebox{\linewidth}{!}{
    \begin{tabular}{p{0.16\linewidth}p{0.84\linewidth}}
        \toprule
        \rowcolor{gray!10}\multicolumn{2}{p{\linewidth}}{
            \begin{minipage}[t]{\linewidth}
                \textbf{[ Dialogue History ]}\\
                \textbf{User}: I'm looking for a hotel to stay in that's expensive and on the east side. \\
                \textbf{Agent}: Express by Holiday Inn Cambridge is on the east side and expensive. \\
                \textbf{User}: That sounds good, but can you tell me if it includes free Wi-Fi?
            \end{minipage}
        } \\
        \midrule

        \textbf{Chosen Response} & 
        \begin{minipage}[t]{0.98\linewidth}
            \textbf{Agent}: \texttt{\textless function\_call\textgreater\{"function" : "find\_hotel"\}\textless/function\_call\textgreater} Yes, it does include free Wi-Fi. Would you like me to book a room for you?
        \end{minipage} \\

        \midrule
        
        \textbf{Rejected Response} & 
        \begin{minipage}[t]{0.98\linewidth}
            \textbf{Agent}: \texttt{\textless function\_call\textgreater\{"function" : "find\_hotel"\}\textless/function\_call\textgreater} Yes, Express by Holiday Inn Cambridge includes free Wi-Fi.
        \end{minipage} \\
        \bottomrule
    \end{tabular}
    }
    \caption{Example of a chosen and rejected response pair for the same dialogue context.}
    \label{tab:appendix_dpo_examples}
\end{table}
\raggedbottom
\newpage
\section{Experimental Setup Details}\label{sec:appendix_experiment}

\subsection{Environments} \label{sec:appendix_env}
All model training was conducted in PyTorch with CUDA 12.1 and Python 3.10, employing two A100 GPUs (80GB each). 
Inference was performed with a combination of two RTX 3090 GPUs (24GB each) and two A100 GPUs (80GB each), utilizing the vLLM\cite{kwon2023efficient} inference framework.

Table~\ref{tab:appendix_model_version} summarizes the LLMs used throughout our experiments, along with their version or release information.

\begin{table}[htbp]
    \centering
    \small
    \begin{tabular}{cc}
        \toprule
        \rowcolor{gray!10} \textbf{Model} & \textbf{Version or Release} \\
        \midrule
        \textsc{Gemini-2.5-Pro} & exp-03-25 \\
        \textsc{GPT‑4o-mini}    & 2024-07-18 \\
        \textsc{GPT-4o}         & 2024-08-06 \\
        \midrule
        \textsc{Llama-3.1-8b-Instruct} & 2024-07-23 \\
        \textsc{Llama-3.2-3b-Instruct} & 2024-09-18 \\
        \textsc{Qwen3-8b} & 2025-05-19 \\
        \bottomrule
    \end{tabular}
    \caption{Versions and release dates of the LLMs used in our experiments.}
    \label{tab:appendix_model_version}
\end{table}

\raggedbottom
\FloatBarrier

\subsection{Evaluation Metrics} \label{sec:appendix_metrics}
We evaluate model performance using three categories of metrics: 
(i) standard TOD metrics (TOD-centric Metrics),
(ii) our proposed flow-aware metrics for proactive mode control (Flow-Aware Metrics), and
(iii) dialogue response quality metrics to assess chitchat responses (Response Quality Metric).

Let $M$ denote the total number of dialogues, and let $T_m$ be the number of turns in dialogue $m$.
We define $\mathcal{D}_m = \{1, \dots, T_m\}$ as the index set of turns in dialogue $m$.
\FloatBarrier

\subsubsection{TOD-centric Metrics}
At each turn $t \in \mathcal{D}_m$, the dialogue has a gold mode $s_t \in \{\text{TOD}, \text{Chitchat}\}$ and a predicted mode $\hat{s}_t$. 
If $s_t = \text{TOD}$, then $I_t$ and $\hat{I}_t$ denote the gold and predicted intents, respectively.
\paragraph{(1) Mode Selection Accuracy and F1-score}
\begin{equation}
\text{Mode Selection Accuracy} = \frac{1}{\sum_{m=1}^{M} T_m} \sum_{m=1}^{M} \sum_{t \in \mathcal{D}_m} \mathbbm{1}[\hat{s}_t = s_t]
\end{equation}

\begin{equation}
\text{Mode Selection F1-score} = \text{Macro-F1 across mode labels (TOD vs. Chitchat)}
\end{equation}





\paragraph{(2) Intent Detection Accuracy}  
We evaluate intent detection accuracy at both the turn and dialogue levels.
Turn-level accuracy measures whether predicted intents are correct on TOD turns(i.e., $\hat{s}_t = \text{TOD}$), while dialogue-level accuracy requires all TOD predictions within a dialogue to be correct. 
Let $M_{\text{TOD}}$ be the number of dialogues in which the model predicts at least one TOD turn. 
Let $\mathcal{D}_m^{\text{TOD-pred}}$ be the set of such turns in dialogue $m$.



\begin{equation}
\text{Intent Accuracy}_\text{/turn} = 
\frac{
    \sum_{m=1}^{M} \sum_{t \in \mathcal{D}_m} 
    \mathbbm{1}[\hat{s}_t = \text{TOD} \wedge \hat{I}_t = I_t]
}{
    \sum_{m=1}^{M} \sum_{t \in \mathcal{D}_m} 
    \mathbbm{1}[\hat{s}_t = \text{TOD}]
}
\end{equation}


\begin{equation}
    \text{Intent Accuracy}_\text{/dialogue} = \frac{1}{M_{\text{TOD}}} \sum_{m=1}^{M} 
\mathbbm{1}[\forall t \in \mathcal{D}_m^{\text{TOD-pred}},\ \hat{I}_t = I_t]
\end{equation}

\paragraph{(3) Joint Accuracy}  
We evaluate whether the model correctly predicts both the mode and, if the mode is TOD, the task intent.

\begin{equation}
\text{Joint Accuracy}_\text{/turn} = 
\frac{1}{\sum_{m=1}^{M} T_m} \sum_{m=1}^{M} \sum_{t \in \mathcal{D}_m} 
\begin{cases}
    \mathbbm{1}[\hat{s}_t = s_t \wedge \hat{I}_t = I_t], & \text{if } s_t = \text{TOD} \\
    \mathbbm{1}[\hat{s}_t = s_t], & \text{if } s_t = \text{Chitchat}
\end{cases}
\end{equation}

\begin{equation}
\text{Joint Accuracy}_\text{/dialogue} = 
\frac{1}{M} \sum_{m=1}^{M} 
\mathbbm{1}\left[
\forall t \in \mathcal{D}_m:
\begin{cases}
    \hat{s}_t = s_t \wedge \hat{I}_t = I_t, & \text{if } s_t = \text{TOD} \\
    \hat{s}_t = s_t, & \text{if } s_t = \text{Chitchat}
\end{cases}
\right]
\end{equation}
\raggedbottom

\subsubsection{Flow-Aware Metrics}
\textit{Switch} and \textit{Recovery} are designed to evaluate whether the model can proactively control dialogue flow across modes. 
Let $\tau^{\text{switch}}_t$ and $\tau^{\text{recovery}}_t$ denote binary indicators of whether the model attempts a switch or recovery at turn $t$. 
Similarly, let $\sigma^{\text{switch}}_t$ and $\sigma^{\text{recovery}}_t$ denote whether the attempt is contextually successful, based on the user's subsequent response.

Switch is computed over all dialogues ($M$), while Recovery is computed only over dialogues where recovery is applicable ($M_{\text{recovery}}$). 
Each value represents the average number of agent-initiated events per dialogue.

\FloatBarrier

\paragraph{(1) Switch}
\begin{equation}
\text{Switch Attempt} = 
\frac{1}{M} \sum_{m=1}^{M} \sum_{t \in \mathcal{D}_m} 
\mathbbm{1}[\tau_t^{\text{switch}} = 1]
\end{equation}

\begin{equation}
\text{Switch Success} = 
\frac{1}{M} \sum_{m=1}^{M} \sum_{t \in \mathcal{D}_m} 
\mathbbm{1}[\sigma_t^{\text{switch}} = 1]
\end{equation}

\raggedbottom

\paragraph{(2) Recovery}
\begin{equation}
\text{Recovery Attempt} = 
\frac{1}{M_{\text{recovery}}} \sum_{m \in \mathcal{M}_{\text{recovery}}} \sum_{t \in \mathcal{D}_m} 
\mathbbm{1}[\tau_t^{\text{recovery}} = 1]
\end{equation}

\begin{equation}
\text{Recovery Success} = 
\frac{1}{M_{\text{recovery}}} \sum_{m \in \mathcal{M}_{\text{recovery}}} \sum_{t \in \mathcal{D}_m} 
\mathbbm{1}[\sigma_t^{\text{recovery}} = 1]
\end{equation}

\subsubsection{Response Quality Metrics and Judging Procedure}
\label{sec:appendix_response_quality}
We define a structured evaluation metric to assess the quality of chat responses in multi-turn dialogue contexts.
The evaluation relies on a pairwise comparison framework, wherein two candidate responses are rated using four qualitative dimensions:

\begin{itemize}
\item \textbf{Sensibleness} — whether the response is coherent and contextually appropriate.
\item \textbf{Specificity} — whether the response is tailored to the preceding dialogue, avoiding generic replies.
\item \textbf{Interestingness} — whether the response is engaging or likely to capture user interest.
\item \textbf{Transition Naturalness} — whether any shift between chitchat and task-oriented dialogue (ToD) occurs smoothly and logically.
\end{itemize}

\begin{figure}[H]
    \centering
    \includegraphics[scale=0.95]{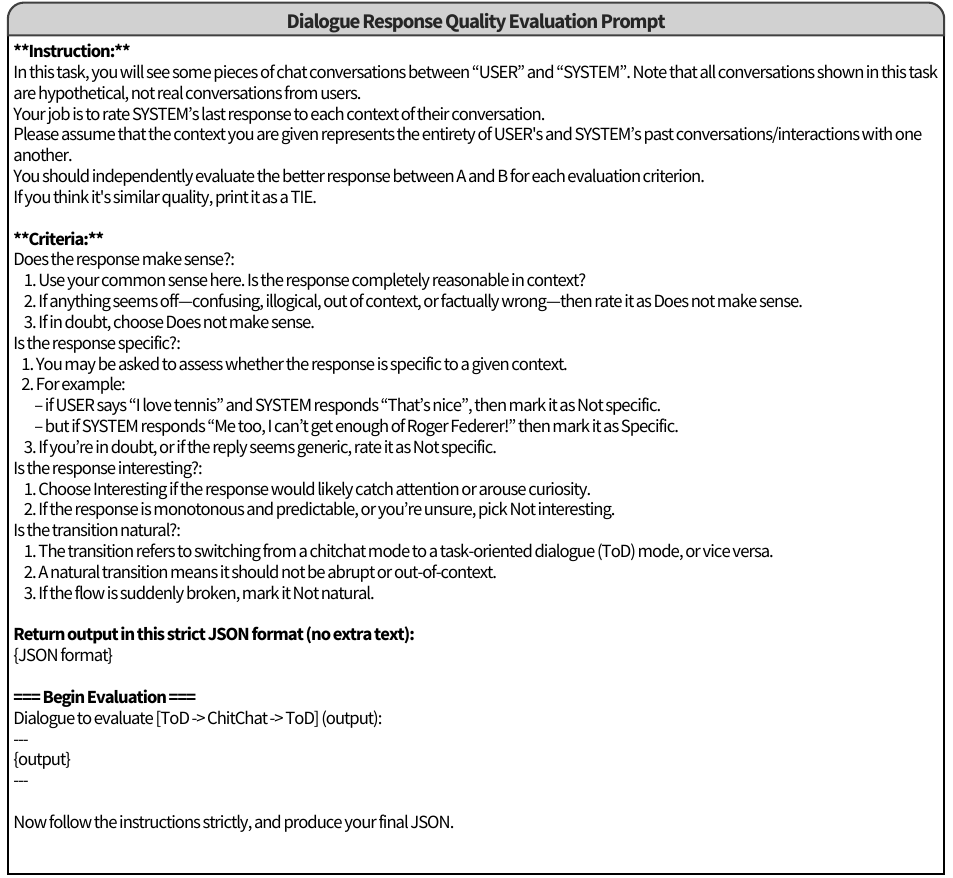}
    \caption{
    Prompt used by the automatic judge (\textsc{Gemini-2.5-Pro}) for dialogue response quality evaluation. 
    This prompt guided pairwise comparisons and was used to evaluate outputs from FnCTOD and \textsc{GPT-4o}, as described in Section~\ref{subsec:qualitative_analysis}.
    }
    \label{fig:prompt_winrate}
\end{figure}

Each comparison is conducted either by an automatic judge (\textsc{Gemini-2.5-Pro}) or by human annotators, following the same evaluation protocol. 
The judge outputs a \texttt{Win/Tie/Lose} decision for each criterion and a concise justification, all formatted as a strict JSON object. 
These outputs are aggregated into per-criterion turn-level win rates, as reported in Section~\ref{subsec:qualitative_analysis}. 
Figure~\ref{fig:prompt_winrate} shows the full evaluation prompt used for the LLM judge, including detailed definitions and instructions to ensure consistency across evaluations. 

\begin{figure}[H]
    \centering
    \includegraphics[scale=0.95]{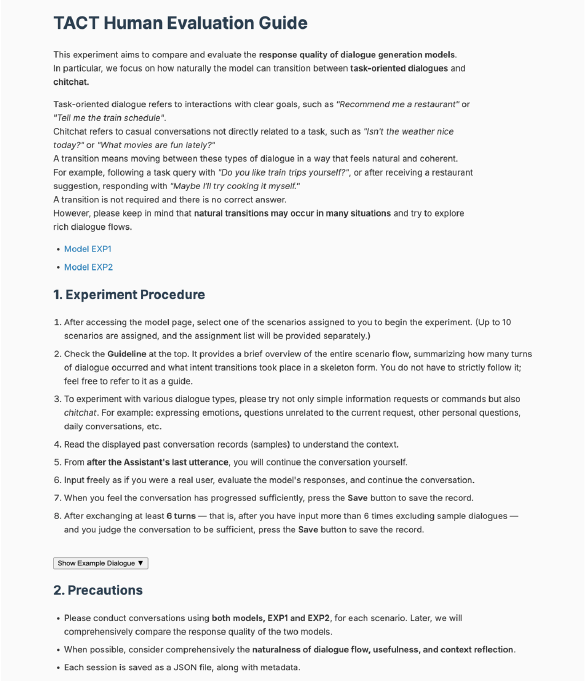}
    \caption{
    Guidelines provided to human annotators for conducting dialogue evaluations. 
    }
    \label{fig:gui_guide}
\end{figure}

\begin{figure}[H]
    \centering
    \includegraphics[scale=0.95]{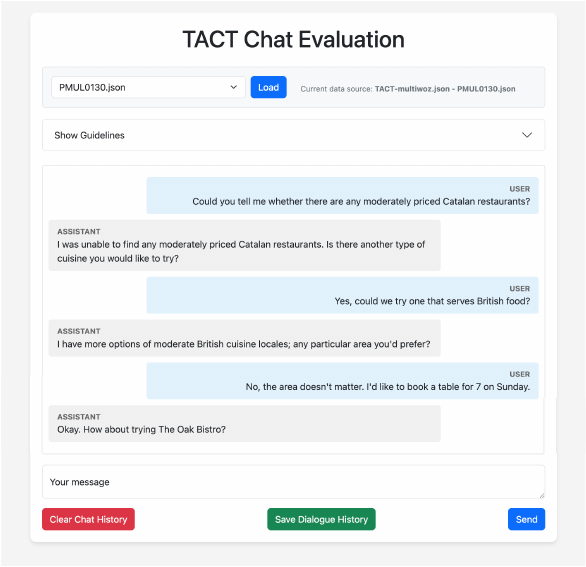}
    \caption{
    Human evaluation interface, where annotators interact with model outputs and record judgments. 
    }
    \label{fig:gui_interface}
\end{figure}

For human evaluation, annotators were provided with a short guideline and a dedicated web interface. 
Figure~\ref{fig:gui_guide} presents the evaluation guidelines, while Figure~\ref{fig:gui_interface} shows the interface used to record judgments. 
The full instruction document and implementation code are available at \url{https://github.com/HYU-NLP/TACT}.

\raggedbottom

\FloatBarrier
\section{Supplementary Results}\label{sec:appendix_discussion}

\subsection{Preliminary Results}
\label{sec:appendix_preliminary_eval}
FnCTOD, trained on $\text{TACT}_{\text{MultiWOZ}}$, showed the strongest overall performance across all metrics, as observed in Table~\ref{tab:performance_sales}.
It achieved the highest accuracy in mode selection and also performed best in turn-level intent detection, making it the most appropriate choice as the final SFT-based baseline. 
Importantly, intent detection scores are computed \textbf{only when the model first correctly classifies the user utterance as task-oriented}.

FnCTOD, trained on $\text{TACT}_{\text{MultiWOZ}}$, showed the strongest overall performance across all metrics, as observed in Table~\ref{tab:performance_sales}.
The corresponding metric formulas are provided in Appendix~\ref{sec:appendix_metrics} for reference.

\begin{table}[h]
    \centering
    \small
    \begin{tabular}{l l l l l}
        \toprule
        \textbf{Dataset} & \textbf{Method} & \textbf{Mode Selection} & \textbf{ID/turn} & \textbf{ID/dialogue} \\
        \midrule
        \multirow{4}{*}{${\text{TACT}_{\text{MultiWOZ}}}$}
            & BERT-base classifier & 77.45 & 96.32 & \textbf{88.61} \\
            & SimpleTOD & 98.86 & 95.39 & 77.50 \\
            & SalesAgent & 93.17 & 82.98 & 34.28 \\
            & \textbf{FnCTOD} & \textbf{98.95} & \textbf{96.35} & 80.94 \\
        \midrule
        \multirow{2}{*}{${\text{TACT}_{\text{SLURP}}}$}     & BERT-base classifier & 72.77 & 93.06 & 80.88 \\
            & \textbf{FnCTOD} & \textbf{97.70} & 92.30 & 78.66 \\
        \bottomrule
    \end{tabular}
    \caption{
    Performance comparison between methods across \tact\ TACT datasets.
    All metrics are accuracy (\%). 
    }
    \label{tab:performance_sales}
\end{table}


\raggedbottom
\subsection{Backbone Comparison} \label{sec:appendix_models}
We further validate whether the comparative trends across methods hold consistently across different backbone models. 
Table \ref{tab:method_comparison_across_models} presents results with \textsc{LLaMA-3.1-8B}, \textsc{LLaMA-3.2-3B}, and \textsc{Qwen-3B}. 
Despite variations in model family and size, we observe the same overall pattern: SFT improves over prompting, and DPO further strengthens response quality metrics (win-rate) while maintaining strong task-oriented performance. 
This suggests that our findings in \S  \ref{sec:quantitative_results} are not specific to a particular architecture, but rather generalize across model choices.

\begin{table}[H]
    \small
    \centering
    \resizebox{\textwidth}{!}{
    \begin{tabular}{caaccaaccaac}
        \toprule
        & \multicolumn{6}{>{\cellcolor{blue!10}}c}{\textbf{TOD}} & \multicolumn{4}{>{\cellcolor{yellow!30}}c}{\textbf{Flow}} & \cellcolor{green!10}\textbf{Chitchat} \\
        \arrayrulecolor{gray!70} \cmidrule(){2-12} \arrayrulecolor{black}
        & \multicolumn{2}{a}{\textbf{Mode Selection}} & \multicolumn{2}{c}{\textbf{Intent Detection}} & \multicolumn{2}{a}{\textbf{Joint Accuracy}} & \multicolumn{2}{c}{\textbf{Switch}} & \multicolumn{2}{a}{\textbf{Recovery}} & \textbf{Overall} \\
        $\textbf{Method}$ & \textbf{Acc.} & \textbf{F1-score} & \textbf{Acc.}/turn & \textbf{Acc.}/dialogue & \textbf{Acc.}/turn & \textbf{Acc.}/dialogue & \textbf{Attempt} & \textbf{Success} & \textbf{Attempt} & \textbf{Success} & \textbf{Win-Rate} \\
        \midrule
        \multicolumn{12}{c}{\textbf{\textsc{Llama-3.1-8b-Instruct}}}\\
        \midrule
        SFT & \textbf{98.95} & \textbf{98.50} & \textbf{96.35} & \textbf{80.94} & \textbf{96.41} & 75.59 & 1.322 & 1.300 & 0.977 & 0.856 & 23.16 \\
        SFT-DPO & 98.82 & 98.32 & 96.03 & 80.00 & 96.21 & \textbf{75.74} & 1.343& \textbf{1.322} & 0.977 & \textbf{0.859} & \textbf{40.86} \\
        \midrule
        \multicolumn{12}{c}{\textbf{\textsc{Llama-3.2-3b-Instruct}}}\\
        \midrule
        SFT & \textbf{99.20} & \textbf{98.89} & \textbf{96.52} & \textbf{82.34} & \textbf{96.98} & \textbf{80.08} & 1.357 & \textbf{1.316} & 0.989 & \textbf{0.449} & 30.23 \\ 
        SFT-DPO & 99.16 & 98.84 & 96.43 & 81.92 & 96.91 & 79.52 & \textbf{1.360} & 1.312 & \textbf{0.992} & 0.446 & \textbf{46.20} \\ 
        \midrule
        \multicolumn{12}{c}{\textbf{\textsc{Qwen3-8b}}}\\
        \midrule
        SFT & 99.29 & 99.02 & \textbf{96.45} & \textbf{82.20} & \textbf{96.97} & \textbf{79.94} & 1.345 & 1.304 & 0.990 & \textbf{0.455} & 33.58 \\ 
        SFT-DPO & \textbf{99.32} & \textbf{99.07} & 96.23 & 81.07 & 96.87 & 79.38 & \textbf{1.383} & \textbf{1.336} & \textbf{0.994} & 0.436 & \textbf{42.56} \\ 
        \bottomrule
    \end{tabular}
    }
    \caption{
    Validation on different backbones (\textsc{LLaMA-3.1-8B}, \textsc{LLaMA-3.2-3B}, \textsc{Qwen-3B}), confirming consistent improvements of DPO over SFT.
    Within each backbone, column-wise best scores are highlighted in \textbf{bold}.
    }
    \label{tab:method_comparison_across_models}
\end{table}
\subsection{$\text{TACT}_\text{SLURP}$ Analysis}\label{sec:appendix_slurp}
Table \ref{table:table_slurp} reports the performance of different methods on TACT\textsubscript{SLURP}.
We observe that SFT achieves consistently higher Intent Detection accuracy compared to ICL baselines, and shows particular strength on flow-related actions such as Switch and Recover. 
Among the pipelines, the classification-based method performs well for per-dialog Intent Detection, but contributes little to mode switching and recovery. 
These results suggest that generation-based approaches are better suited for real-world applications where smooth switching and recovery between conversation modes are critical.


\begin{table}[H]
    \small
    \centering
    \resizebox{\textwidth}{!}{%
    \begin{tabular}{caaccaaccaa}
        \toprule
        & \multicolumn{6}{>{\cellcolor{blue!10}}c}{\textbf{TOD}} & \multicolumn{4}{>{\cellcolor{yellow!30}}c}{\textbf{Flow}}  \\
        \arrayrulecolor{gray!70} \cmidrule(){2-11} \arrayrulecolor{black}
        \textbf{ } & \multicolumn{2}{a}{\textbf{Mode Selection}} & \multicolumn{2}{c}{\textbf{Intent Detection}} & \multicolumn{2}{a}{\textbf{Joint Accuracy}} &\multicolumn{2}{c}{\textbf{Switch}} & \multicolumn{2}{a}{\textbf{Recover}} \\ 
        $\textbf{Method}$ &  \textbf{Acc.} & \textbf{F1-score} & \textbf{Acc.}/turn & \textbf{Acc.}/dialogue & \textbf{Acc.}/turn & \textbf{Acc.}/dialogue & \textbf{attempt} & \textbf{success} & \textbf{attempt} & \textbf{success} \\ 
        \midrule
        ICL-ZS & 88.89 & 88.80 & 84.15 & 61.76 & 82.49 & 25.88 & 0.973 & 0.406 & \textbf{1.417} & 0.308 \\ 
        ICL-FS & 93.67 & 93.65 & 85.79 & 64.85 & 88.25 & 43.97 & 0.710 & 0.433 & 0.717 & 0.360 \\ 
        \midrule
        SFT & \textbf{97.70} & \textbf{97.70} & \textbf{92.30} & \textbf{78.66} & \textbf{94.86} & \textbf{72.46} & \textbf{1.478} & \textbf{0.883} & 1.105 & \textbf{0.373} \\
        \bottomrule 
    \end{tabular}%
    }
    \caption{
    Results of different methods on TACT\textsubscript{SLURP}. 
    The best score in each column is \textbf{bolded}.
    }
    \label{table:table_slurp}
\end{table}
\raggedbottom
\subsection{Examples of Transition Handling}
\label{sec:appendix_transition_samples}

Table~\ref{tab:appendix_model_comparison} provides a qualitative example of model responses in a dialogue following the TCT pattern. 
In this dialogue, the user temporarily shifts from booking a train to discussing their enjoyment of train journeys, allowing us to observe how different models manage the conversational transition. 
SFT ignores the chitchat and resumes the task abruptly, indicating a lack of sensitivity to conversational flow.
In contrast, \textsc{GPT-4o} (ICL-FS) acknowledges the user’s sentiment but fails to provide a concrete next step in the task, showing weakness in task grounding. 
DPO, however, handles the transition most naturally: it validates the user’s comment and then smoothly returns to the original task with a relevant train recommendation, demonstrating both contextual awareness and task alignment. 
This example highlights the effectiveness of DPO in enabling more human-like, fluid transitions between modes.




\begin{table}[h]
    \centering
    \small
    \renewcommand{\arraystretch}{1}
    \resizebox{\linewidth}{!}{
    \begin{tabular}{p{0.17\linewidth} p{0.83\linewidth}} 
        \toprule
        \rowcolor{gray!10}\multicolumn{2}{p{\linewidth}}{ 
        \begin{minipage}[t]{\linewidth}
            \textbf{[ Dialogue History ]} \\
            \textbf{User}: Can you help me find a train? I'll be traveling on Wednesday. \\
            \textbf{Agent}: What is your departure point and what time would you like to depart? \\
            \textbf{User}: Oh, of course. Sorry. I'm leaving out of Leicester and going into Cambridge. \\
            \textbf{Agent}: Is there a specific time you want to leave or arrive by? I have 2,828 trains to pick from. I need to narrow this down. \\
            \textbf{User}: I want to arrive by 16:15. \\
            \textbf{Agent}: That narrowed it down to 10 options. What time would you like to leave? \\
            \textbf{User}: I don't have a preference on leaving time. By the way, I really enjoy train journeys. There's something so relaxing about watching the scenery go by. Do you have a favorite route? \\
            \textbf{Agent}: Oh, I completely agree! Train journeys can be quite scenic. The route from Leicester to Cambridge has some lovely views, especially as you approach the countryside. \\
            \textbf{User}: That sounds nice! I love seeing fields and little villages.
            It makes the trip feel more like an adventure.
        \end{minipage}
        } \\
        \midrule
        \textbf{SFT Response} & 
        \begin{minipage}[t]{\linewidth}
            \textbf{Agent}: How about TR8076? It leaves at 05:09. \\
            \textcolor{red}{(Ignores the chitchat and returns directly to task.)}
        \end{minipage} \\
        \midrule
        \textbf{SFT-DPO Response} & 
        \begin{minipage}[t]{\linewidth}
            \textbf{Agent}: Absolutely! It adds a nice touch to the journey. Now, back to your train options. \\ I would recommend TR 7853. It leaves at 13:09 and arrives at 14:54. \\
            \textcolor{blue}{(Most natural transition—acknowledges the chitchat and resumes the task smoothly.)}
        \end{minipage} \\
        \midrule
        \textbf{ICL-FS Response} &
        \begin{minipage}[t]{\linewidth}
            \textbf{Agent}: Absolutely, the countryside views are charming! Now, let's finalize your train booking. \\ Would you like me to proceed with one of the options that arrive by 16:15? \\
            \textcolor{red}{(Acknowledges the chitchat but lacks concrete task grounding.)}
        \end{minipage} \\
        \bottomrule
    \end{tabular}
    }
    \caption{Responses from different models given the same dialogue.
    DPO better acknowledges the user’s input and naturally recovers to the task.}
    \label{tab:appendix_model_comparison}
\end{table}

\section*{Terms for Use of Artifacts}
Our work builds upon two publicly available benchmarks: MultiWOZ~(MIT license) and SLURP~(CC BY 4.0 license). 
The proposed TACT dataset, constructed by augmenting these corpora, will be released under the CC BY 4.0 license. 
All resources are intended for academic and non-commercial use, and proper attribution is required when using them.

\end{document}